\documentclass[12pt]{article}
\pdfoutput=1
\usepackage[utf8]{inputenc}
\usepackage[T1]{fontenc}
\usepackage{float}
\usepackage{authblk}
\usepackage[normalem]{ulem}
\usepackage{amsmath}
\usepackage{amsfonts}
\usepackage{amsthm}
\usepackage{adjustbox}
\usepackage{graphicx}
\usepackage{booktabs}
\usepackage{mathtools}
\usepackage{fullpage}
\usepackage[mathlines]{lineno}
\usepackage{subcaption}
\usepackage[style=numeric-comp,sorting=none,maxnames=25]{biblatex} 
\usepackage{hanging}
\usepackage{longtable}
\usepackage{algorithm}
\usepackage{algorithmic}

\usepackage{todonotes}

\usepackage{multirow}

\usepackage[colorlinks = true,
            linkcolor = black,
            urlcolor  = black,
            citecolor = teal]{hyperref}

\title{Supervised structure learning}

\author[1,2]{Karl~J.~Friston}
\author[1,2,3]{Lancelot Da Costa}
\author[2]{Alexander Tschantz}
\author[2]{Alex Kiefer}
\author[2]{Tommaso Salvatori}
\author[1]{Victorita Neacsu}
\author[2]{Magnus Koudahl}
\author[2]{Conor Heins}
\author[1]{Noor Sajid}
\author[5]{Dimitrije Markovic}
\author[4]{Thomas Parr}
\author[2]{Tim Verbelen}
\author[2]{Christopher L Buckley}
\affil[1]{Wellcome Trust Centre for Neuroimaging, Institute of Neurology, University College London, UK.}
\affil[2]{VERSES AI Research Lab, Los Angeles, California, 90016, USA}
\affil[3]{Department of Mathematics, Imperial College London, UK}
\affil[4]{Nuffield Department of Clinical Neurosciences, University of Oxford, UK}
\affil[5]{Chair of Cognitive Computational Neuroscience, Technische Universit\"at Dresden, Dresden, Germany}

\begin{document}
\maketitle
\pagenumbering{arabic}

\begin{abstract}
This paper concerns structure learning or discovery of discrete generative models. It focuses on Bayesian model selection and the assimilation of training data or content, with a special emphasis on the order in which data are ingested. A key move—in the ensuing schemes—is to place priors on the selection of models, based upon expected free energy. In this setting, expected free energy reduces to a constrained mutual information, where the constraints inherit from priors over outcomes (i.e., preferred outcomes). The resulting scheme is first used to perform image classification on the MNIST dataset to illustrate the basic idea, and then tested on a more challenging problem of discovering models with dynamics, using a simple sprite-based visual disentanglement paradigm and the Tower of Hanoi (cf., blocks world) problem. In these examples, generative models are constructed autodidactically to recover (i.e., disentangle) the factorial structure of latent states—and their characteristic paths or dynamics.
\end{abstract}

\newpage

\section{Introduction}\label{sec:introduction}

This paper addresses the problem of automatically discovering or building generative models through Bayesian model selection \cite{cooper1992bayesian, friedman2003being}, under active inference and learning \cite{friston2016activelearning}. Active inference refers to the maximisation of model evidence (a.k.a., marginal likelihood) using variational bounds; namely, variational free energy as a bound on log evidence (a.k.a. an evidence lower bound \cite{winn2005variational}), when the generative model entails the consequences of action. In such a setting, the implicit self-evidencing can be read as gathering evidence for an agent’s model of the world \cite{hohwy2016self, parr2019generalised, ramstead2023bayesian}. The requisite planning as (active) inference \cite{attias2003planning} rests upon prior beliefs about the way data are sampled. These priors are supplied by the free energy expected on committing to a particular action or policy \cite{parr2019generalised}. Here, we generalise the notion of action to include selection, not just of training data, but of the generative models that best explain those data. In brief, this leads to three levels of belief updating that can be ascribed to \textit{inference}, \textit{learning} and \textit{selection}. Here, these nested optimisation processes correspond to Bayesian belief updating about latent \textit{states}, model \textit{parameters} and \textit{structure}, respectively. The focus of this work is on model selection or structure learning, of the kind that would normally be addressed using nonparametric Bayes or variants of manifold learning: e.g., \cite{zemel2004proximity, gershman2012tutorial}. By restricting ourselves to discrete state-space models, we simplify this problem to deciding: do we augment the model—given this new observation—or not?

In the context of structure learning, Bayesian model selection considers two models, one with and one without an extra component (e.g., an additional latent state) and evaluates the marginal likelihood of both, as scored by their variational free energy for the same (new) observation. If both models were equally likely, \textit{a priori}, then the marginal likelihoods become posteriors over the two models, and the model with the greatest posterior probability would be selected to explain the next observation—and so on, until the model is sufficiently expressive to explain all new data. However, this presupposes that the new data were generated by some latent states that have and have not been encountered previously, with equal probability. Clearly, this prior is not appropriate when assimilating or ingesting data, because the probability that the current data is a novel exemplar decreases with the number of exemplars or training data previously encountered. In short, one needs a prior over the relative probabilities of the model with and without a new latent state. Perhaps the most established priors of this sort are those found in nonparametric Bayes, usually based on stick-breaking processes that inherit from certain assumptions about distributions from which data are sampled; e.g., \cite{teh2006}. 

However, there is another constraint or prior over models, which follows from the application of the free energy principle to structure learning. This prior arises when treating model selection as an active process that minimises expected free energy. Intuitively, in the absence of constraints on prior expectations over outcomes, (negative) expected free energy reduces to the mutual information between (unobservable) latent states and (observable) content generated by those states \cite{parr2019generalised}. This means one can supplement the differences in variational free energy (i.e., log marginal likelihoods) with differences in mutual information (as log model priors) to score the posterior probability of an augmented model, relative to the original model. In principle, the model will grow, with new exemplars, until there are no more training exemplars that require a novel explanation—and the mutual information asymptotes.

We can extend this line of thinking to any structural aspect of a generative model. This becomes relatively straightforward in the case of discrete state-space models, as exemplified by partially observed Markov decision processes (POMDPs). In the simplest case of a hidden Markov model, the model selection described above just entails evaluating the posterior probability of models with and without an additional latent state, having optimised the parameters of both models with respect to variational free energy. To underwrite the generality of this (active) model selection, we will use an expressive generative model—that is hierarchically composable—in which hidden \textit{states} factorise under some mean field approximation. Furthermore, we generalise the usual POMDP and represent \textit{paths} or trajectories as latent variables. In a conventional Markov decision process, state transitions are conditioned upon an action or a control variable. However, this variable can be read as a path that may or may not be actionable (i.e., may or may not be controllable)\footnote{The implicit generalisation can be likened to the use of generalised coordinates in physics, where \textit{position} and \textit{momentum} are treated as distinct random variables. On this view, action can be regarded as a force that changes the path through state-space.}. 

With this sort of generative model, one can select from a space of models that include: (i) the original or parent model, (ii) a model with an additional state in each factor; (iii) a model with an additional path within each factor or (iv) a new factor. A new factor can only have two states and one path, where the first path is the simplest; namely, a stationary path on which each state transitions to itself. The ensuing models can be scored in terms of their variational free energy and the most likely model selected, under suitable model priors. This can be repeated for successive epochs of data, until the model stops growing. There are two key issues that this formulation of structure learning brings to the table.

\subsection*{The importance of order}

First, the order in which data are encountered—or presented, i.e. the scheduling or curriculum \cite{al2011evolutionary, bengio2009curriculum}—matters. This follows from the fact that we are dealing with state-space models that have to learn the transitions or dynamics, in terms of a discrete number of paths. The key theme here is that the scheduling or curriculum under which the data are presented \cite{al2011evolutionary, bengio2009curriculum}. In other words, learning of the dynamics or physics is only possible if data are presented in the order in which they are generated. This means that there is some requisite supervision of structure learning; in the sense that the process generating training data has to respect their ordinal structure. Clearly, this is not an issue if the data are generated by the process being learned. However, it suggests that it is not possible to do structure learning or any form of disentanglement \cite{higgins2021unsupervised, sanchez2020learning} in the absence of ordinal structure. As we will see later, this applies even in the context of static recognition or classification, where the kind of structure learning implied here depends upon serially uncorrelated sampling from training data.

\subsection*{The importance of being discrete}

The second issue is the commitment to discrete state-space models. This speaks to the secondary agenda of this paper; namely, to foreground the utility of discrete state-space models in relation to the (implicit) generative models used in most of deep learning. Here, we are reading deep learning as synonymous with the use of backpropagation of errors and requisite differentiability \cite{lecun2015deep}. Differentiability restricts models to continuous state-spaces of the kind that support embedding. Continuous state-space models offer many attractive features; especially, in high-dimensional state-spaces. However, there are certain things one can do in discrete state-space models that elude continuous formulations \cite{da2020active, fields2023control2, fields2022free, wauthier2022learning,dacostaHowActiveInference2022a}. For example:

\begin{enumerate}
    \item Explainability is assured by equipping discrete states (and paths) with an explicit human-interpretable semantics (and syntax).
    \item The functional form of posteriors and priors (i.e., categorical and Dirichlet distributions) admit flexible and polymodal probabilistic representations.
    \item Deep architectures and nonlinearities are replaced by a single likelihood tensor, mapping from latent states to discrete outcomes.
    \item It is straightforward to compute expected information gain (i.e., expected free energy) in a discrete setting, using tensor operators and the linear algebra found in quantum information theory.
    \item Finally, it is straightforward to encode uncertainty about model parameters in the form of Dirichlet distributions and accompanying tensor computations.
\end{enumerate}

The last point is particularly important from the perspective of active inference—and possibly for all biomimetic schemes \cite{fields2023control}. This follows because it is impossible to optimise posterior beliefs about the latent causes of content without inferring model parameters, and \textit{vice versa}. This means that optimal inference and learning rests upon probability distributions \textit{over model parameters} (i.e., tensors in discrete state-spaces and connection weights in continuous models). This issue is prescient in the context of structure learning, because Bayesian model selection requires posterior beliefs over both latent states and parameters\footnote{To evaluate (bounds on) marginal likelihood (a.k.a., model evidence), one has to marginalise over parameter posteriors.}, which require probability distributions over connection weights in conventional architectures (e.g., Bayesian deep learning). These considerations licence our focus on discrete state-space models, which may offer a complement to continuous state-space models.

This paper is organised as follows. First, we provide a brief but complete description of the generative model and its variational inversion, with a special focus on active inference, learning and selection predicated on expected free energy. We then illustrate some of the basic notions behind structure learning—as active selection—by application to a machine learning classification benchmark (MNIST digits) \cite{bach2015pixel}. We then consider the structure learning of state-space models and the learning of dynamics in Euclidean spaces from pixels, using a setup analogous to the dSprites dataset \cite{dsprites} but simplified  \cite{champion2022branching, jeon2021ib}. Finally, we apply the same variational procedures to solve the Tower of Hanoi problem that would usually be addressed using techniques such as Planning Domain Definition Languages \cite{fox2003pddl2}. We conclude with a discussion of the ensuing procedures and ask to what extent they complement conventional approaches based upon nonparametric Bayes and continuous state-space formulations.

\section{Active inference}\label{sec:active_inference}

In this section, we rehearse the structure of the models used in the numerical studies of subsequent sections. This model can be regarded as a generalisation of a partially observed Markov decision process (POMDP). The generalisation in question covers trajectories, narratives or syntax—which may or may not be controllable—by equipping a POMDP with random variables called \textit{paths}. Paths effectively pick out dynamics or transitions among latent states. These models are designed to be composed hierarchically, in a way that speaks to a separation of temporal scales in deep generative models. In other words, the number of transitions among latent states at any given level is greater than the number of transitions at the level above. This furnishes a unique specification of a hierarchy, in which the parents of any latent \textit{factor} (associated with unique states and paths) contextualise the dynamics of their children.

The variational inference scheme \cite{beal2003variational} used to invert these models inherits from their application to online decision-making tasks. This means that action selection rests primarily on current beliefs about latent states and structures, and expectations about future observations. In that sense, posterior beliefs are updated sequentially in an online fashion with each new action-outcome pair. This calls for Bayesian filtering (i.e., forward message passing) during the active sampling of observations, followed by Bayesian smoothing (i.e., forward and backward message passing) to revise posterior beliefs about past states at the end of an epoch. Bayesian smoothing ensures that the beliefs about latent states at any moment in the past are informed by all available observations when updating model parameters (and latent states of parents in deep models). 

In neurobiology, this combination of Bayesian filtering and smoothing would correspond to evidence accumulation during active engagement with the environment, followed by a ‘replay’ before the next epoch \cite{buckner2010role, louie2001temporally, penny2013forward, pezzulo2014internally}. From a machine learning perspective, this can be regarded as a forward pass (c.f., belief propagation) for online active inference, followed by a backwards pass (implemented with variational message passing) for active learning. For completeness, we have tried to put all the requisite expressions for (forward) message passing in figures and accompanying legends. The implicit belief updates, pertaining to states, parameters and structure, foreground the interdependencies between active inference, learning, and selection, respectively.

\subsection*{Generative modelling}

Active inference rests upon a \textit{generative model} of observable outcomes (observations). This model is used to infer the most likely causes of outcomes in terms of expected states of the world. These states (and paths) are latent or \textit{hidden} because they can only be inferred through observations. Some paths are controllable in the sense they can be realised through action. Therefore, certain observations depend upon action (e.g., where one is looking), which requires the generative model to entertain expectations about outcomes under different combinations of actions (i.e., policies)\footnote{Note that in this setting, a policy is not a sequence of actions but simply a combination of paths, where each hidden factor has an associated state and path. This means there are, potentially, as many policies as there are combinations of paths.}. These expectations are optimised by minimising \textit{variational free energy}. Crucially, the prior probability of a policy depends upon its \textit{expected free energy}. Having evaluated the expected free energy of each policy, the most likely action can be selected and the perception-action cycle continues \cite{parr2022active}.

\subsection*{The generative model}

\begin{figure*}[t!]
    \centering
    \includegraphics[width=0.5\textwidth]{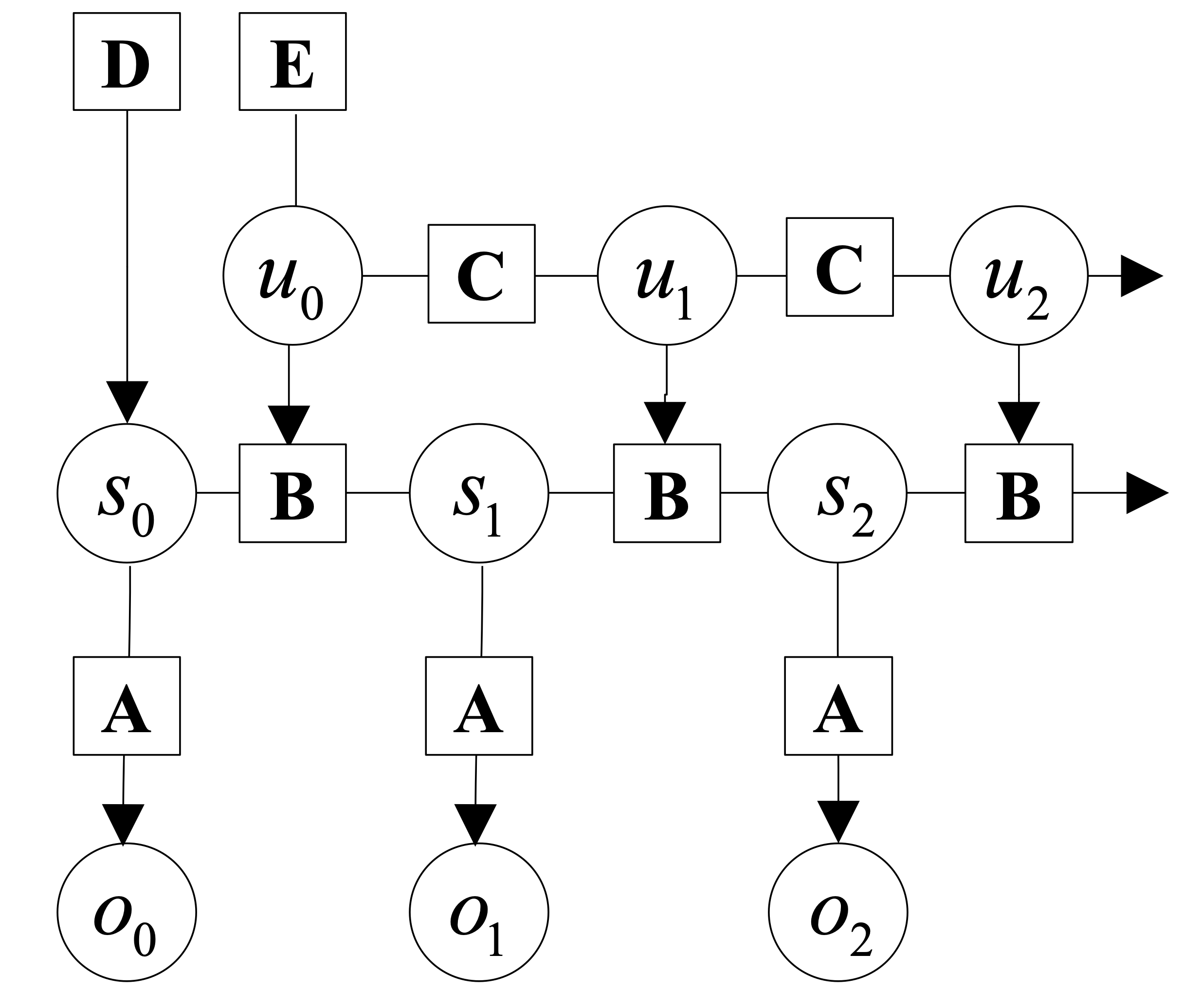}
    \caption{\textbf{Generative models as agents.} A generative model specifies the joint probability of observable consequences and their hidden causes. Usually, the model is expressed in terms of a \textit{likelihood} (the probability of consequences given their causes) and \textit{priors} (over causes). When a prior depends upon a random variable it is called an \textit{empirical prior}. Here, the likelihood is specified by a tensor $\mathbf{A}$, encoding the probability of an outcome under every combination of \textit{states} ($s$). The empirical priors pertain to transitions among hidden states, $\mathbf{B}$, that depend upon \textit{paths} ($u$), whose transition probabilities are encoded in $\mathbf{C}$. $\mathbf{E}$ specifies the empirical prior probability of each path.}
\label{fig:figure_1b}
\end{figure*}

Figure \ref{fig:figure_1b} provides a schematic overview of the generative model used for the problems considered in this paper. Outcomes at any particular time depend upon hidden \textit{states}, while transitions among hidden states depend upon \textit{paths}. Note that paths are random variables, in the sense that a particle can have both a position (i.e., a state) and momentum (i.e., a path). Paths may or may not depend upon action. The resulting POMDP is specified by a set of tensors. The first set of parameters, denoted $\mathbf{A}$, maps from hidden states to outcome modalities; for example, exteroceptive (e.g., visual) or proprioceptive (e.g., eye position) \textit{modalities}. These parameters encode the likelihood of an outcome given their hidden causes. The second set $\mathbf{B}$ prescribes transitions among the hidden states of a \textit{factor}, under a particular path. Factors correspond to different kinds of causes; e.g., the location versus the class of an object. The remaining tensors encode prior beliefs about paths $\mathbf{C}$, and initial states $\mathbf{D}$. The tensors—encoding probabilistic mappings or contingencies—are generally parameterised as Dirichlet distributions, whose sufficient statistics are concentration parameters or \textit{Dirichlet counts}. These count the number of times a particular combination of states or outcomes has been inferred. We will focus on learning the likelihood model, encoded by Dirichlet counts, $\boldsymbol{a}$.

The generative model in Figure \ref{fig:figure_1b} means that outcomes are generated as follows: first, a policy is selected using a softmax function of expected free energy. Sequences of hidden states are generated using the probability transitions specified by the selected combination of paths (i.e., policy). Finally, these hidden states generate outcomes in one or more modalities. Perception or inference about hidden states (i.e., state estimation) corresponds to inverting a generative model, given a sequence of outcomes, while learning corresponds to updating model parameters. Perception therefore corresponds to updating beliefs about hidden states and paths, while learning corresponds to accumulating knowledge in the form of Dirichlet counts. The requisite expectations constitute the sufficient statistics $(\mathbf{s}, \mathbf{u}, \mathbf{a})$ of posterior beliefs $Q(s, u, a) = Q_{\mathbf{s}}(s)Q_{\mathbf{u}}(u)Q_{\mathbf{a}}(a)$. The implicit factorisation of this approximate posterior effectively partitions model inversion into inference, planning and learning.

\subsection*{Variational free energy and inference}

In variational Bayesian inference (a form of approximate Bayesian inference), model inversion entails the minimisation of variational free energy with respect to the sufficient statistics of approximate posterior beliefs. This can be expressed as follows, where, for clarity, we will deal with a single factor, such that the policy (i.e., combination of paths) becomes the path, $\pi = u$. Omitting dependencies on previous states, we have for model $m$:

\begin{equation}
\begin{aligned}
Q\left(s_\tau, u_\tau, a\right) & =\arg \min _Q F \\
F & =\mathbb{E}_Q[\ln \underbrace{Q\left(s_\tau, u_\tau, a\right)}_{\text {posterior }}-\ln \underbrace{P\left(o_\tau \mid s_\tau, u_\tau, a\right)}_{\text {likelihood }}-\ln \underbrace{P \left(s_\tau, u_\tau, a\right)}_{\text {prior }}] \\
& =\underbrace{D_{K L}\left[Q\left(s_\tau, u_\tau, a\right) \| P\left(s_\tau, u_\tau, a \mid o_\tau\right)\right]}_{\text {divergence }}-\underbrace{\ln P\left(o_\tau \mid m\right)}_{\text {log evidence }} \\
& =\underbrace{D_{K L}\left[Q\left(s_\tau, u_\tau, a\right) \| P\left(s_\tau, u_\tau, a\right)\right]}_{\text {complexity }}-\underbrace{\mathbb{E}_Q\left[\ln P\left(o_\tau \mid s_\tau, u_\tau, a\right)\right]}_{\text {accuracy }}
\label{eq:1}
\end{aligned}
\end{equation}

Because the (KL) divergences cannot be less than zero, the penultimate equality means that free energy is zero when the (approximate) posterior is the true posterior. At this point, the free energy becomes the negative log evidence for the generative model \cite{beal2003variational}. This means minimising free energy is equivalent to maximising model evidence, which is equivalent to minimising the complexity of accurate explanations for observed outcomes.

Planning emerges under active inference by placing priors over (controllable) paths to minimise expected free energy \cite{friston2015active}:

\begin{equation}
\begin{aligned}
& G(u)=\mathbb{E}_{Q_{u}}\left[\ln Q\left(s_{\tau+1}, a \mid u\right)-\ln Q\left(s_{\tau+1}, a \mid o_{\tau+1}, u\right)-\ln P\left(o_{\tau+1} \mid c\right)\right] \\ \\
& =-\underbrace{\mathbb{E}_{Q_{u}}\left[\ln Q\left(a \mid s_{\tau+1}, o_{\tau+1}, u\right)-\ln Q(a \mid s_{\tau+1}, u)\right]}_{\text {expected information gain (learning) }}- \\
& \underbrace{\mathbb{E}_{Q_{u}}\left[\ln Q\left(s_{\tau+1} \mid o_{\tau+1}, u\right)-\ln Q\left(s_{\tau+1} \mid u\right)\right]}_{\text {expected information gain (inference) }} \underbrace{-\mathbb{E}_{Q_{u}}\left[\ln P\left(o_{\tau+1} \mid c\right)\right]}_{\text {expected cost }} \\ \\
& =-\underbrace{\mathbb{E}_{Q_{u}}\left[D_{K L}\left[Q\left(a \mid s_{\tau+1}, o_{\tau+1}, u\right) \| Q(a \mid s_{\tau+1},u)\right]\right]}_{\text {novelty }}+ \\
& \underbrace{D_{K L}\left[Q\left(o_{\tau+1} \mid u\right) \| P\left(o_{\tau+1} \mid c\right)\right]}_{\text {risk }} \underbrace{-\mathbb{E}_{Q_{u}}\left[\ln Q\left(o_{\tau+1} \mid s_{\tau+1}, u\right)\right]}_{\text {ambiguity }} \\
&
\label{eq:2}
\end{aligned}
\end{equation}

Here, 
\begin{multline*}
 Q_u=Q\left(o_{\tau+1}, s_{\tau+1}, a \mid u\right) = P\left(o_{\tau+1}, s_{\tau+1}, a \mid u, o_0, \ldots, o_\tau\right) = 
 P\left(o_{\tau+1} \mid s_{\tau+1}, a\right) Q\left(s_{\tau+1}, a \mid u\right)   
\end{multline*}
is the posterior predictive distribution over parameters, hidden states and outcomes at the next time step, under a particular path. Note that the expectation is over \textit{observations in the future} that become random variables; hence, \textit{expected} free energy. This means that preferred outcomes—that subtend expected cost and risk—are prior beliefs, which constrain the implicit planning as inference 
\cite{attias2003planning, botvinick2012planning, van2013informational}.

One can also express the prior over the parameters in terms of an expected free energy, where, marginalising over paths:

\begin{equation}
\begin{aligned}
P(a) & =\sigma(-G) \\
G(a) & =\mathbb{E}_{Q_a}[\ln P(s \mid a)-\ln P(s \mid o, a)-\ln P(o \mid c)] \\
& =-\underbrace{\mathbb{E}_{Q_a}[\ln P(s \mid o, a)-\ln P(s \mid a)]}_{\text {expected information gain}} -\underbrace{\mathbb{E}_{Q_a}[\ln P(o \mid c)]}_{\text {expected value}} \\
& =-\underbrace{\mathbb{E}_{Q_a}\left[D_{K L}[P(o, s \mid a) \| P(o \mid a) P(s \mid a)]\right.}_{\text {mutual information}} \underbrace{-\mathbb{E}_{Q_a}[\ln P(o \mid c)]}_{\text {expected cost}}
\label{eq:3}
\end{aligned}
\end{equation}

where $Q_a = P(o|s, a) P(s|a) = P(o, s|a)$ is the joint distribution over outcomes and hidden states, encoded by the Dirichlet parameters, $a$, and $\sigma(\cdot)$ is the softmax function. Note that the Dirichlet parameters encode the mutual information, in the sense that they implicitly encode the joint distribution over outcomes and their hidden causes. When normalising each column of the $a$ tensor, we recover the likelihood distribution (as in Figure \ref{fig:figure_1}); however, we could normalise over every element, to recover a joint distribution.

Expected free energy can be regarded as a universal objective function that augments mutual information with expected costs or constraints. Constraints —parameterised by $c$— reflect the fact that we are dealing with open systems with characteristic outcomes, $a$. This can be read as an expression of the constrained maximum entropy principle that is dual to the free energy principle \cite{ramstead2023bayesian}. Alternatively, it can be read as a constrained principle of maximum mutual information or minimum redundancy \cite{ay2008predictive, barlow1961possible, linsker1990perceptual, olshausen1996emergence}. In machine learning, this kind of objective function underwrites disentanglement \cite{higgins2021unsupervised, sanchez2020learning}, and generally leads to sparse representations \cite{gros2009cognitive, olshausen1996emergence, sakthivadivel2022weak, tipping2001sparse}.

When comparing the expressions for expected free energy in (\ref{eq:2}) with variational free energy in (\ref{eq:1}), the expected divergence becomes expected information gain. Expected information gain about the parameters and states are sometimes associated with distinct epistemic affordances; namely, \textit{novelty} and \textit{salience}, respectively \cite{schwartenbeck2019computational}. Similarly, expected log evidence becomes expected value, where value is the logarithm of prior preferences. The last equality in (\ref{eq:2}) provides a complementary interpretation; in which the expected complexity becomes risk, while expected inaccuracy becomes ambiguity.

There are many special cases of minimising expected free energy. For example, maximising expected information gain maximises (expected) Bayesian surprise \cite{itti2009bayesian}, in accord with the principles of optimal (Bayesian) experimental design \cite{lindley1956measure}.This resolution of uncertainty is related to artificial curiosity \cite{schmidhuber1991curious, still2012information} and speaks to the value of information \cite{howard1966information}.

Expected complexity or risk is the same quantity minimised in risk sensitive or KL control \cite{klyubin2005empowerment, broek2012risk}, and underpins (free energy) formulations of bounded rationality based on complexity costs \cite{braun2011path, ortega2013thermodynamics} and related schemes in machine learning; e.g., Bayesian reinforcement learning \cite{ghavamzadeh2015bayesian}. More generally, minimising expected cost subsumes Bayesian decision theory \cite{berger2013statistical}.

This class of generative models can be further expanded with hierarchical and factorial depth, which makes it a core building block of universal generative models. Moreover, because of subscribing to discrete state and observation spaces, the variational message passing scheme becomes remarkably simple and corresponds to a fixed point iteration scheme. More details are provided in Appendix~\ref{app:ugm}.

\section{Active selection}\label{sec:active_selection}

In contrast to learning—that optimises \textit{posteriors} over parameters—Bayesian model selection or structure learning \cite{tenenbaum2011grow, tervo2016toward, tomasello2016cultural} can be framed as optimising the \textit{priors} over model parameters. Bayesian model reduction is a top-down approach to this kind of structure learning, which starts with an expressive model and removes redundant parameters.\footnote{A relevant example is the training of ``clone-structured cognitive graphs'' based on a combination of E-M and Viterbi training \cite{georgeetal2021}. Here, a hidden Markov model with a restricted likelihood structure is learned, such that after training most Dirichlet counts on an initially random transition matrix are set to 0, resulting in a sparse graph that recovers the structure of the target domain. The inductive bias embodied in the clone structure can be regarded as a specific prior over the likelihood mapping, while the Viterbi training induces an additional sparsity bias on the transition matrix parameters.} Crucially, Bayesian model reduction can be applied to the posterior beliefs after the data have been assimilated. In other words, Bayesian model reduction is a \textit{post hoc} optimisation that refines current beliefs based upon alternative models that may provide potentially simpler explanations \cite{friston2011post}.

Bayesian model reduction is a generalisation of ubiquitous procedures in statistics \cite{savage1954foundations}. In the present setting, it reduces to something remarkably simple: by applying Bayes rules to parent and reduced models it is straightforward to show that the change in free energy can be expressed in terms of posterior Dirichlet counts $\mathbf{a}$, prior counts $a$ and the prior counts that define a reduced model $a^{\prime}$. Using $\mathcal {B}$ to denote the beta function, we have \cite{friston2018bayesian}:

\begin{equation}
\begin{aligned}
\Delta F & =\ln P(o \mid a)-\ln P\left(o \mid a^{\prime}\right) \\
& =\ln \mathcal{B}(\mathbf{a})+\ln \mathcal{B}\left(a^{\prime}\right)-\ln \mathcal{B}(a)-\ln \mathcal{B}\left(\mathbf{a}+a^{\prime}-a\right) \\
\mathbf{a}^{\prime} & =\mathbf{a}+a^{\prime}-a
\end{aligned}
\end{equation}

Here, $\mathbf{a}^{\prime}$ corresponds to the posterior that one would have obtained under the reduced priors. Please see \cite{friston2020second, neacsu2022structure, smith2020active}, for worked examples in epidemiology and neuroscience.

The alternative to Bayesian model reduction is the bottom-up expansion of models to accommodate new data or content. If one considers the selection of one (parent) model over another (augmented) model as an action, then the difference in expected free energy furnishes a log prior over models that can be combined with the (variational free energy bound on) log marginal likelihoods to score their posterior probability. This can be expressed in terms of a log Bayes factor (i.e., odds ratio) comparing the likelihood of two models, given some observation, $o$:

\begin{equation}
\begin{aligned}
\ln \frac{P(m \mid o)}{P\left(m^{\prime} \mid o\right)} & =\ln \frac{P(o \mid m) P(m)}{P\left(o \mid m^{\prime}\right) P\left(m^{\prime}\right)}=\Delta F+\Delta G \\
\Delta F & =\ln P(o \mid m)-\ln P\left(o \mid m^{\prime}\right) \\
\Delta G & =\ln P(m)-\ln P\left(m^{\prime}\right)=G(\mathbf{a} \mid m)-G\left(\mathbf{a}^{\prime} \mid m^{\prime}\right)
\label{eq:8}
\end{aligned}
\end{equation}

Here, $\mathbf{a}$ and $\mathbf{a}^{\prime}$ denote the posterior expectations of parameters under a parent $m$, and augmented model $m^{\prime}$, respectively. The difference in expected free energy reflects the information gain in selecting one model over the other, following (\ref{eq:3})\footnote{Expected free energy additionally scores the degree to which some model accounts for \textit{preferred} outcomes (see Equation \ref{eq:3}). Hence, models are more plausible \textit{a priori} if they can explain portions of state space that align with an agent's preferences or goals \cite{tschantz2020learning}.}. One can now retain or reject the parent model, depending upon whether the log odds ratio is greater than or less than zero, respectively. This kind of (active) model selection therefore finds structures with precise or unambiguous likelihood mappings. 

In the special case of structure learning, one is effectively testing the hypothesis that each outcome\footnote{Generally specified probabilistically.} is generated either by a previously unseen state of affairs or has been previously encountered. This means that each outcome is generated by a particular combination of latent states (and paths). In turn, this requires Bayesian model selection to consider models under precise priors over latent states (and paths). An augmented model thereby assumes the outcome is generated by a new state (or path), while the parent model assumes the outcome is generated by the most likely state (or path) previously experienced. 

The structure learning illustrated below uses the variational free energy of states (and paths) to select among the parent and augmented models. If an augmented model has greater evidence, it is only accepted if the expected free energy affords an improvement over the parent model, with the same priors over model parameters.  Another way of looking at this procedure, is that models with greater model evidence are compressed in a way that minimises information loss. In what follows, we apply these procedures to some familiar test cases.

\begin{algorithm}
\caption{Model comparison procedure}
\begin{algorithmic} 

\STATE $\Delta F_{\textrm{min}} \gets 0$
\STATE $m_{\textrm{best}} \gets m$
\WHILE{other structures to compare}
    \STATE $m^{\prime} \gets \textrm{\textbf{expand\_model}}(m)$
    \STATE \COMMENT{Compare variational free energy (has marginal likelihood increased?)}
    \STATE $\Delta F \gets \ln P(o \mid m)-\ln P\left(o \mid m^{\prime}\right)$ 
    \IF{ $\Delta F <  \Delta F_{\textrm{min}}$  } 
      \STATE \COMMENT{Compare expected free energy (has mutual information increased?)}
      \STATE $\Delta G \gets \ln P(m) - \ln P(m^{\prime})$
          \IF{ $\Delta G >  0 $ }
                \STATE $\Delta F_{\textrm{min}} \gets \Delta F $
                \STATE $m_{\textrm{best}} \gets m^{\prime}$
        \ENDIF
        \ENDIF
\ENDWHILE
\RETURN $m_{\textrm{best}}$
\end{algorithmic}
\end{algorithm}
\section{MNIST revisited}\label{sec:mnist_revisited}

In this section, we use a numerical example to illustrate the active learning and selection afforded by minimising expected free energy. We focus on a familiar image classification problem; namely the MNIST digits \cite{bach2015pixel, scellier2017equilibrium}, approaching this problem from the perspective of discrete state-space modelling. The classification or recognition problem at hand is an inference problem; namely, inferring the most likely latent state generating an observed collection of pixels. In the absence of a generative model, we have to select the most likely model structure and learn the most likely parameters within that structure. For the simple case of recognising static patterns, the generative model is a tensor mapping from latent states to outcomes. Posterior beliefs about the parameters of this likelihood tensor can be encoded with Dirichlet parameters that, effectively, count the co-occurrences of a particular combination of states and outcomes. The log likelihood of an observation, given a combination of latent states, is then given by digamma functions of the Dirichlet counts: $
\mathbb{E}\left[\ln P\left(\mathbf{A}^g\right)\right]=\varphi\left(a^g\right)$. 

Note that because we are dealing with a static model, there is only one (stationary) path and the transition tensors ($\mathbf{B}$) reduce to identity matrices. Note further, that we can use the conditional independence of the outcomes, given latent states, to parameterise the likelihood ($\mathbf{A}$) with a separate tensor for each outcome modality: 

\begin{equation}
\begin{aligned}
\ln P\left(o_\tau^1, \ldots, o_\tau^G \mid s_\tau, a\right) & =\sum_g \ln P\left(o_\tau^g \mid s_\tau, a\right) \Rightarrow \\
\mathbb{E}_Q\left[\ln P\left(o_\tau^1, \ldots, o_\tau^G \mid s_\tau, a\right)\right] & =\sum_{g \in \operatorname{ch}(f)} o_\tau^g \odot \varphi(a^g) \odot_{i \in pa(g) \backslash f} s_\tau^i   
\end{aligned}
\end{equation}

In short, instead of dealing with a large image, the discrete formulation deals with lots of little tensors. For example, in the MNIST dataset, images are a collection of (28×28) 784 pixels, each pixel representing an outcome modality. Each modality can have a number of levels. In the case of the MNIST digits, they can be black or white. Or, in the case of supplying observations probabilistically, the probability of being black or white. The likelihood tensors used here deal with categorical outcomes (i.e., black pixel or white pixel). However, the pixel values themselves, following normalisation, are continuous numbers between zero and one, which can be treated as probabilities of a pixel being black or white. Treating pixel intensities as (probabilities of) categorical outcomes greatly simplifies the ensuing inferences. This just leaves the problem of enumerating the latent states generating the outcomes at each pixel. So, what do we know about digits? We know that there are at least two factors. In order to generate a digit we need to know its class (between ‘0’ and ‘9’). Second, we need to know the ‘style’ in which it has been written. Style here encompasses every way in which one could write a particular digit. Clearly, there can be one hundred and one ways of writing a digit. So if we commit to this prior belief, can we learn the generative model from scratch?

In this demonstration of active learning and selection, we know the class label of each digit, which means one can place a precise prior over the states of the digit factor. However, we do not know any digit’s style or, indeed, how many styles are apt to generate handwritten digits. However, by accumulating evidence from successive exemplars—sampled at random from any given digit class—we can start with a minimal generative model, in which there is just one style, and use that to explain the first exemplar\footnote{We initialise the first (and subsequent) columns of the likelihood tensor with a symmetric Dirichlet distribution parameterised with a small concentration parameter. In these examples, we use 1/16.}. Practically, this just means adding the (probabilistic) observation to the one and only column of the likelihood tensor in the form of Dirichlet parameters (see Figure \ref{fig:figure_1b}). We can then take the second (training) exemplar and ask whether an additional style state is licensed by comparing the evidence for models with one and two latent style states, suitably augmented with priors over the two models. 

The expected free energy supplies a generic prior in terms of the increase or decrease in the mutual information\footnote{Note that the (negative) expected free energy of the parameters reduces to the mutual information because there is no preferred outcome in these classification problems.} afforded by adding an extra column to the likelihood tensor. Because this is effectively a log prior, we can now augment this with priors based upon our prior beliefs about styles. If there are $N$ styles, then the probability that the current exemplar has seen before is simply the proportion of the total number of styles previously encountered (denoted below as $H$)\footnote{ Clearly, this is a naïve Bayesian prior. One could consider the priors used in nonparametric Bayes that assume an infinite number of styles or classes \cite{gershman2012tutorial}. Alternatively, one could turn to the statistical literature on species discovery by treating styles as ‘species’ \cite{efron1976estimating, fisher1943relation}. However, the naïve prior above is sufficient for our purposes and produces characteristic style (c.f., species) discovery curves, as we will see later.}. This simple hyperprior can now be used to evaluate the posterior odds of the parent model, relative to a model with an additional (unseen) style.

\begin{equation}
\begin{aligned} p & =\frac{n}{N}, \quad a \in \mathbb{R}_{+}^{\ell \times n} \\ \ln \frac{P(m \mid o)}{P\left(m^{\prime} \mid o\right)} & =\Delta F+\Delta G+\Delta H \\ \Delta H & =\ell(p)-\ell(1-p)\end{aligned}
\end{equation}

One can now accept or reject the model with a second latent state and proceed to the third exemplar, and so on. One can then repeat this for every digit class to assemble or learn the requisite likelihood tensor for every pixel. One could even repeat this entire procedure under different hyperpriors about the upper bound on the number of styles. In the examples below, we assumed that there could be 128 styles. This hyperprior means that to be 99.9\% certain we have encountered every style at least once (assuming all styles are equiprobable) we would need about 1000 training examples: i.e., $\ell(0.001) / \ell\left(1-\frac{1}{N}\right) \approx 880$.

\begin{figure*}[t!]
    \centering
\includegraphics[width=0.6\textwidth]{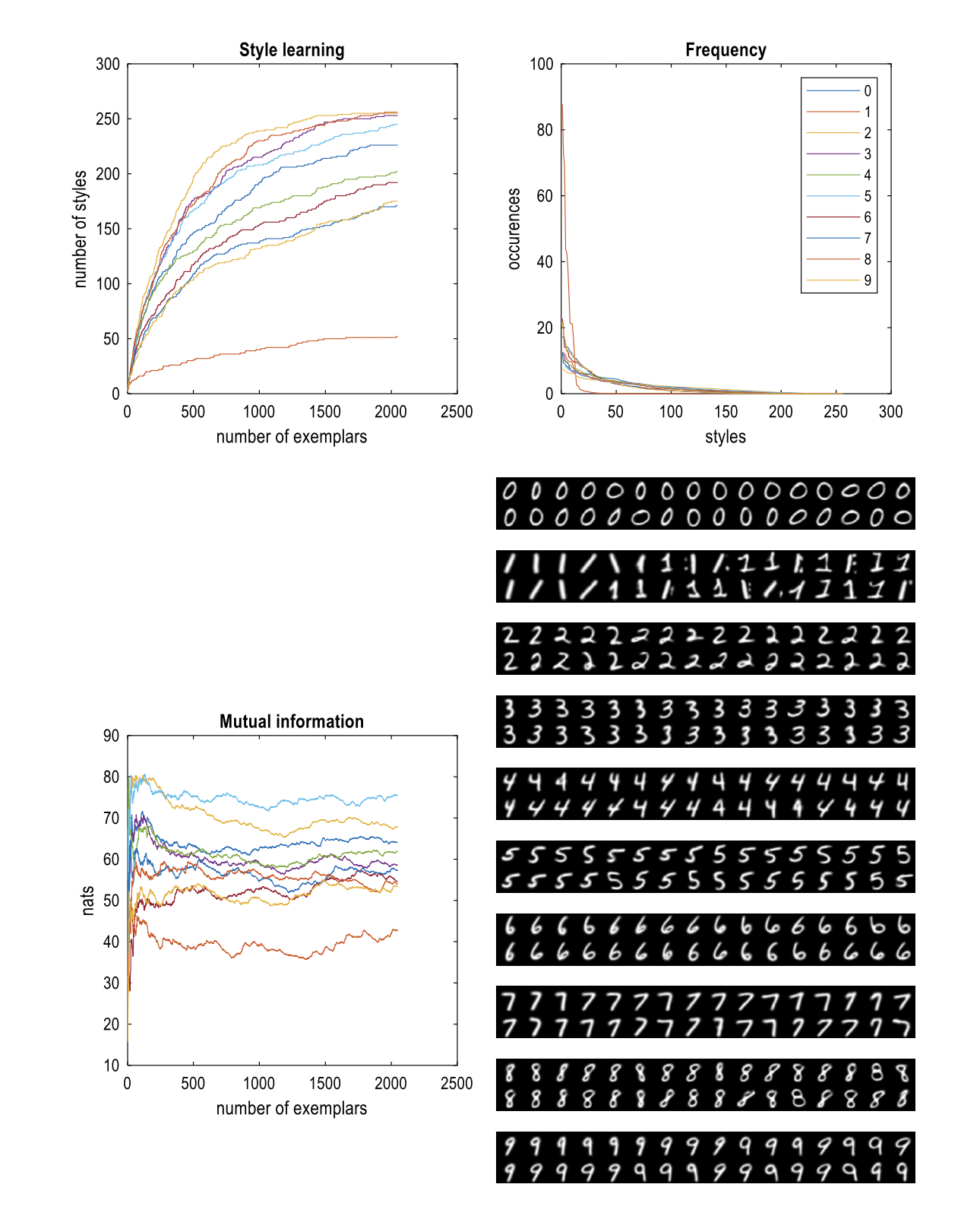}
\caption{\textbf{
Style learning:
}  This figure reports the results of structure learning of the MNIST training data. The upper right panel shows the number of styles learned as a function of the number of training exemplars. The coloured curves correspond to the digit classes.  These discovery curves differ markedly among the numbers. For example, the number “8” appears to have many more styles (red line) than the number “1” (orange line). Note how the rate of discovery decreases with the number of exemplars, as one would expect from style or species discovery curves \cite{efron1976estimating, fisher1943relation}. The upper right panel shows the number of instances of each digit class over the discovered styles. This has a roughly exponential distribution for most numbers with the exception of the number “1”—that is dominated by relatively few (about 16) distinct styles. The remaining styles are very rare. The lower left panel shows the mutual information (i.e., negative expected free energy) associated with the likelihood mapping during the ingestion of exemplars. This rises quickly during the initial exemplars and is then maintained at a relatively high level. This is because a new style is only accepted if there is no loss of mutual information (note that updating the Dirichlet counts of a previously seen style can reduce the mutual information). Interestingly, one can see that the most informative number is “5”, while the least informative is number “1”. The lower right panel illustrates the first 32 styles for each of the 10 digits classes. This is what each style and digit class looks like, in observation space, according to the discovered likelihood mapping. }
\label{fig:figure_3}
\end{figure*}

The results in Figure \ref{fig:figure_3}, were obtained by using the first 2048 training exemplars of each digit class, from the MNIST dataset\footnote{The digits were downloaded from \href{https://lucidar.me/en/matlab/load-mnist-database-of-handwritten-digits-in-matlab/}{https://lucidar.me/en/matlab/load-mnist-database-of-handwritten-digits-in-matlab/} and pre-processed by smoothing with a Gaussian convolution kernel (of two pixels width) and subject to histogram equalisation. The 512 most informative pixels were used for structure learning and subsequent classification.}. The upper left panel shows the total number of styles as a function of the number of exemplars showing that new exemplars are ingested or assimilated more slowly as learning (and selection) proceeds: a characteristic of species discovery curves \cite{efron1976estimating}. Interestingly, there appear to be many fewer styles for the digit “1” than any of the others with the digit “8” having the greatest repertoire of styles. The upper left panel shows the distribution over styles for the 10 digit classes. This suggests that for most digits there are about 50 common styles, with the remaining styles occurring very infrequently; i.e., once or twice. The mutual information increases quickly during the assimilation of new styles at the beginning of learning (and selection) and then declines slowly as more exemplars are installed into the likelihood tensor. As might be anticipated, the digit “1” has the smallest mutual information because it has a more limited repertoire of styles. The lower right panel in Figure 3 shows the first 32 styles. These are automatically ordered, roughly, in terms of their frequency because common styles are accumulated earlier in training, relative to rare styles that are added sporadically later in training.

\subsection*{Classification and inference}

\begin{figure*}[t!]
    \centering
\includegraphics[width=0.7\textwidth]{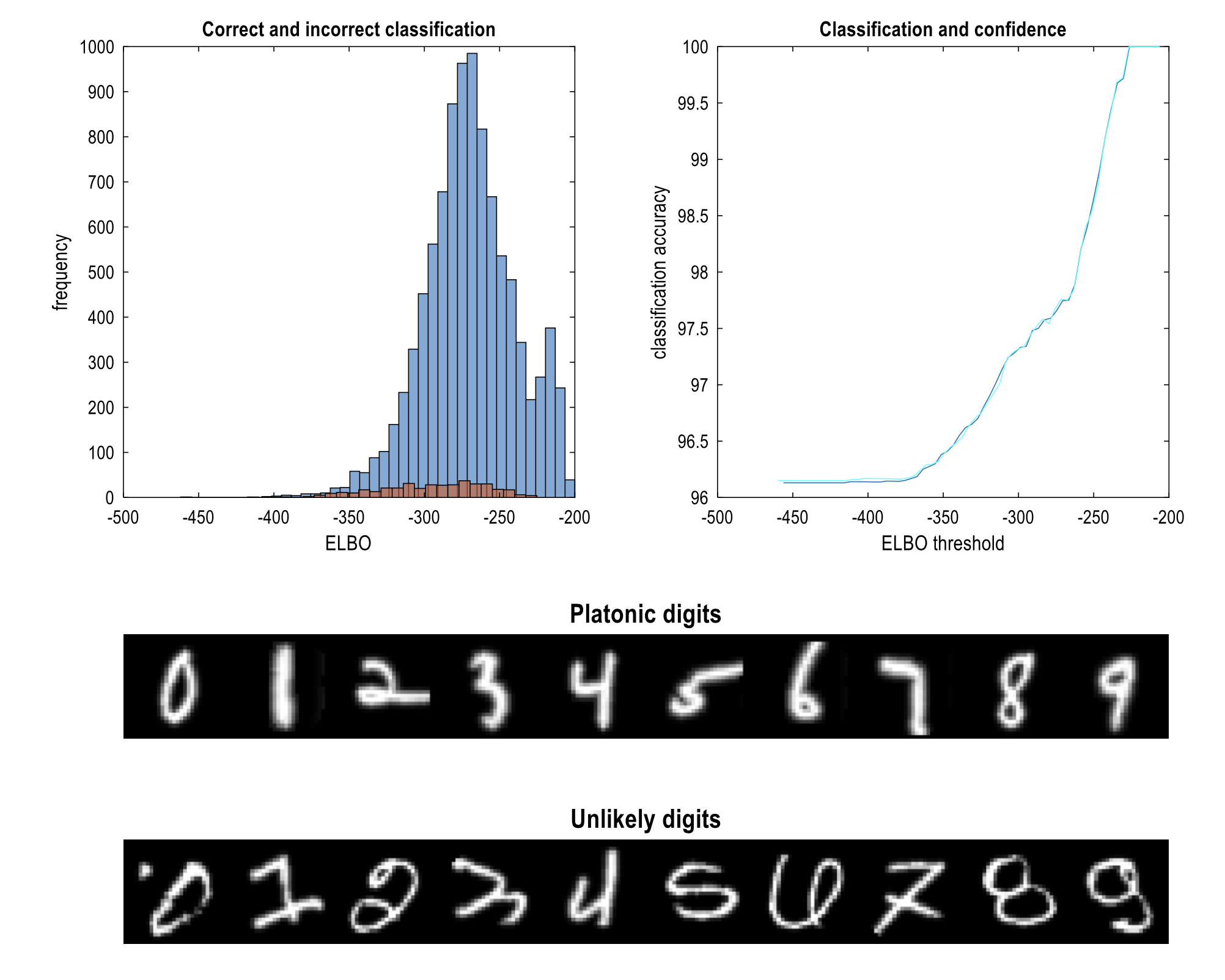}
\caption{\textbf{
Classification performance.
}  The upper left panel illustrates the distribution of log marginal likelihood (as evaluated using the variational free energy or ELBO) over the 20,480 digits ingested or assimilated. The two distributions correspond to classifications that match the supervised class labels (blue – correct, brown – incorrect). The corresponding classification accuracy is shown on the upper right as a function of ELBO threshold. This shows that all digits were classified with an accuracy of over 96\%. This climbs to 100\% for the few thousand or so digits classified with a high marginal likelihood. The lower panel shows the predictions of training exemplars with a high (Platonic) and low marginal likelihood (Unlikely digits). The blue and cyan lines reporting classification accuracy correspond to performance before (blue) and after (cyan) Bayesian model reduction or pruning of redundant likelihood parameters.
 }
\label{fig:figure_4}
\end{figure*}

Figure \ref{fig:figure_4} shows the results of classification performance when inferring the most likely digit using 10,000 test images. The classification accuracy for the entire test cohort is unremarkable in relation to state-of-the-art machine learning (above 99\%\footnote{\href{https://paperswithcode.com/sota/image-classification-on-mnist}{https://paperswithcode.com/sota/image-classification-on-mnist}}), standing at about 96\%. However, variational inference allows one to evaluate accuracy as a function of the confidence that the test image was a recognisable digit.

One can assess the likelihood—-that any given image is a digit—-using the marginal likelihood afforded by the negative variational free energy. The marginal likelihood can be read as the validity of any ensuing classification (from the model's perspective). The upper left panel of Figure \ref{fig:figure_4} plots the sample distribution of log marginal likelihoods for correctly classified images (blue) and incorrectly classified images (brown). The key thing to note is that when digits have a high marginal likelihood they can be classified with high accuracy. This is shown in the upper right that plots the classification accuracy as a function of the threshold applied to triage unlikely images. The images in the lower panels show test images with the highest and lowest marginal likelihoods.

\subsection*{Bayesian model reduction}

The preceding numerical studies illustrate some basic variational procedures. This example of structure learning is semi-supervised in the sense that we knew the latent state corresponding to the digit class; enabling structure learning within each level of this \textit{class} factor. Although classification accuracy is unremarkable, this classification comes equipped with a confidence that the image could be classified as a digit in the first place. Here, this confidence was quantified in terms of the variational bound on (log) marginal likelihood. Crucially, this kind of scheme also quantifies uncertainty about the model parameters, in terms of Dirichlet distributions over the likelihood mapping. This is potentially important because it allows us to further optimise the model with respect to variational free energy using Bayesian model reduction: see \ref{eq:8}. 

For example, we can ask whether model evidence increases or decreases when setting small Dirichlet counts to zero. In other words, having grown our model using structure learning \cite{tenenbaum2011grow}, we can now prune redundant parameters using Bayesian model reduction \cite{hobson2012waking}. The full (selected) model contained 2,336,768 likelihood parameters. After model reduction, we eliminated 65,448 parameters. This increased log marginal likelihood by 5.53 natural units per output modality (i.e., pixel) with a marginal increase in classification accuracy (96.12\% to 96.14\%): see the cyan line in the upper panel of Figure \ref{fig:figure_4}. In principle, this reduced model should generalise slightly better to new test data \cite{sengupta2018robust}.

\subsection*{Information geometry and embedding spaces}

If the above scheme is Bayes optimal, why did it fail to achieve state-of-the-art classification accuracy? One answer to this question is that continuous formulations—of the sort used in deep learning—have clear advantages over discrete formulations\footnote{Another answer is that the discrete state space scheme did not consider a factorial or compositional structure for styles that were conserved over levels of the digit class factor. For example, one might consider a mean field approximation where digits are generated from multiple factors; e.g., digit class, translation, rotation, width of strokes, curvature, et cetera.}. One advantage is that they can express prior constraints on the way that content is generated. An example of these constraints is the prior knowledge that images have certain contiguity properties (e.g., smoothness) that can be leveraged using convolutional neural networks and implicit weight-sharing \cite{lecun2015deep}. Although one could consider similar constraints in the above example—e.g., by sharing Dirichlet parameters between adjacent pixels—this would rely upon some metric structure that is the province of continuous models. In other words, by committing to continuous state-spaces, one induces well-defined metrics that can be used to articulate prior beliefs about the generation of content. For example, images are generated via a projective geometry in a Euclidean space. One can extend this notion to specific problems; for example, digits are generated by composing various line segments in Euclidean space: e.g., \cite{gklezakos2022active}, and so on. In contrast, discrete state-spaces are background-free, in the sense that they do not have an explicit notion of space or time.

However, there is an information geometry at hand in discrete state-space models. This follows from the fact that one can read the parameters of a generative model as sufficient statistics. To illustrate this point, one can regard the columns of tensors as points on a high-dimensional statistical manifold \cite{amari2016information, ay2015information, caticha2015basics, kim2018investigating}. The distances between points on this manifold are given by the information length; namely, the path integral of infinitesimal KL divergences as one moves from the probability distribution encoded by a particular location on the statistical manifold to another\footnote{Strictly speaking, the information length is the path integral of the square root of twice the infinitesimal KL divergence.} \cite{crooks2007measuring, kim2018investigating,dacostaNeuralDynamicsActive2021}. 

\begin{figure*}[t!]
    \centering
\includegraphics[width=0.7\textwidth]{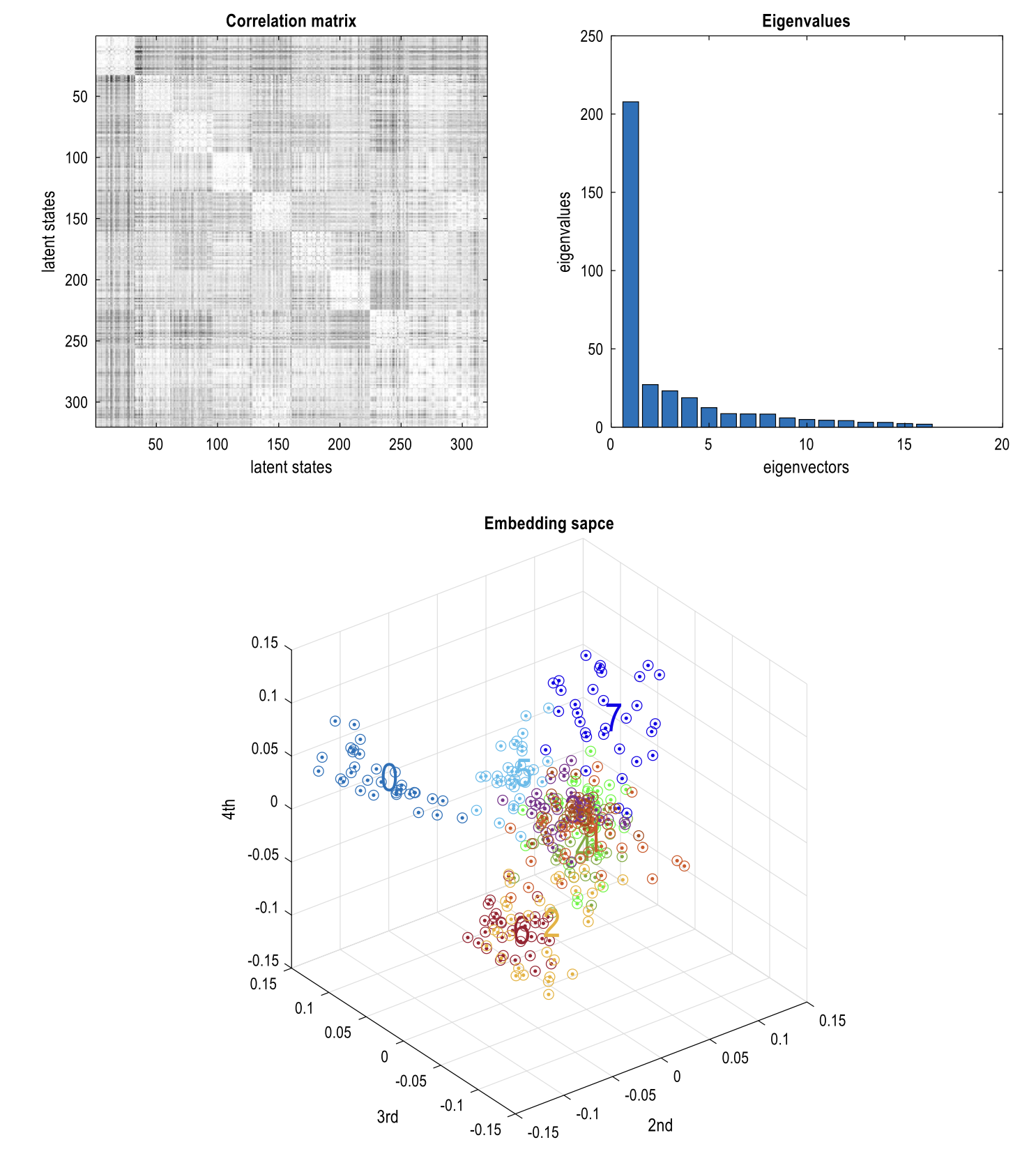}
\caption{\textbf{
Information geometry and embedding spaces.
}  This figure reports the implicit embedding spaces in terms of the information geometry induced by the likelihood mappings. The upper left panel shows a correlation or similarity matrix based upon the Jeffreys divergence between (the first 32) styles of different digit classes. The 10 digit classes can be seen in the block diagonal structure over the 32 styles of each digit (i.e., those shown in Figure 3). The ensuing metric space can be characterised in terms of its eigenvectors using singular value decomposition of the correlation matrix on the upper left. This follows because there always exists a hypersphere in which each latent state occupies a unique position, so that the Euclidean distance to all other latent states corresponds to the Jeffreys divergence between their respective likelihood mappings. One can take many different perspectives on the ensuing space. The lower panel depicts the locations of 32 styles for each of the 10 digit classes in the subspace spanned by the second, third and fourth eigenvectors. This illustrates a separation between the digit class “0” and remaining digits. Similar separations can be seen from other perspectives (i.e., rotations) of this space. The eigenvalues in the upper right panel score the dispersion of latent states on the hypersphere; illustrating that most of the dispersion or metric variance lies in a low dimensional subspace.
 }
\label{fig:figure_5}
\end{figure*}

One can use this high-dimensional statistical manifold to look at the consequence of structure learning. Figure \ref{fig:figure_5} shows the ensuing statistical embedding space defined in terms of information length. Here, we approximated the distance between each combination of latent states (i.e., the categorical distributions encoded in the columns of the $\mathbf{A}$ tensor) using the symmetric (Jeffreys) KL divergence. Averaging (the square of) this metric allows one to create a correlation matrix $\rho$, measuring the similarities (c.f., cosine) between different latent states. This appeals to standard procedures in multidimensional scaling \cite{edelman1998representation}, in which each latent state can be regarded as lying on a hypersphere, whose radius is equal to the maximum distance $D_{i j}$ between latent states $i$ and $j$:

\begin{equation}
\begin{aligned} & \Delta_{i j}^2=2 \cdot\left(1-\rho_{i j}\right) \Rightarrow \rho_{i j}=1-\frac{1}{2} \Delta_{i j}^2 \\ & \Delta_{i j}=2 \cdot \frac{D_{i j}}{\max _{i j} D_{i j}}, D_{i j}^2=\sum_g\left(d_{i j}^g\right)^2 \\ & d_{i j}^g=D_{K L}\left[\mu\left(a_i^g\right) \| \mu\left(a_j^g\right)\right]+D_{K L}\left[\mu\left(a_j^g\right) \| \mu\left(a_i^g\right)\right]\end{aligned}
\end{equation}

The principal eigenvectors of the ensuing correlation matrix (upper left panel) are the principal coordinates of the high dimensional (c.f., embedding) space, over which representations—of particular digits in particular styles—are dispersed. The degree of dispersion is given by the accompanying eigenvalues (upper right panel), suggesting that latent states are largely dispersed in a five dimensional subspace of the statistical manifold. The lower panel of Figure \ref{fig:figure_5} plots the locations of latent states, colour-coded according to digit class. This kind of embedding space is an emergent property of structure learning or discovery; namely, maximising marginal likelihood and mutual information. 

\subsection*{Summary: the importance of being discrete}

The foregoing suggests that one can solve image classification benchmarks in a straightforward way that eschews much of the art of machine learning: namely, there is no need for backpropagation of errors, or  for \textit{ad hoc} cost functions; there is no need to specify a neural network architecture or to choose appropriate nonlinear functions, and so on. Committing to discrete state (and outcome) spaces, means that there exists an optimal likelihood tensor that maps from latent states (e.g., labels) to observable data (e.g., content): cf., \cite{salakhutdinov2012learning}.

Anecdotally, when moving from continuous to discrete or quantised state-spaces, one is effectively replacing deep compositions of nonlinear mappings in low dimensional (continuous) state-spaces with one linear operation in a high dimensional (discrete) state-space. To the extent that this is true, one might ask whether hierarchical discrete state-space models are necessary; for example, hierarchical Dirichlet processes \cite{mackay1995hierarchical, teh2006}? One answer rests upon a separation of temporal scales: when moving from static to dynamic (a.k.a., state-space) models, the game changes and we have to consider the discrete transitions that are apt for any given context. Here, context is constituted by a superordinate state-space that—by definition—unfolds at a slower rate. This is why deep discrete state-space models are generally synonymous with deep temporal models that bring a semi-Markovian aspect to the way that content is generated, e.g., natural language \cite{friston2018deep, george2009towards, mackay1995hierarchical}. In the next section, we tackle the problem of structure learning in the setting of state transitions--that supply priors on the latent states--to supplement the likelihood model considered in this section.

\section{Dynamics and disentanglement}\label{sec:disentangling_dsprites}

In this section, we generalise the procedures of the previous section to accommodate dynamics. The previous example dealt with static images, which is the limiting case of generative models where there is only one (stationary) path encoded by an identity matrix. Here, we address the problem of structure learning from exemplars in which there can be lawful transitions between the states of one or more factors. This leads to more expressive state-space models that foreground the role of the priors encoded in the transition tensors. In this setting, one can grow a model in a number of directions. Starting with a model with a single latent state, for each outcome modality, one can add a state to the first factor. If there are two or more states, one can add a path to the factor, or add a new factor, with two hidden states. Note that adding a factor with two latent states (and one path) is mandated because the first state is shared with other factors. This follows because there are implicitly many other factors with only one state (and stationary path). This leads to the following recipe for model expansion:

\begin{enumerate}
    \item If this is the first observation, create a likelihood mapping with a single state and populate it with initial Dirichlet counts (i.e., the concentration parameter of a symmetric Dirichlet distribution\footnote{Typically, the concentration parameter of a prior (symmetric) Dirichlet distribution is chosen to be less than one; such that samples have a sparse distribution. Anecdotally, the smaller the concentration parameter the more impressionable the model is, when assimilating a new observation. In this section, the concentration parameter was 1/16.}).
    \item For subsequent observations consider an additional state for the last factor, provided there is only one (stationary) path. Otherwise, consider an additional path or factor (with two states and one stationary path).
\end{enumerate}

This protocol simplifies the space of models considered; in that once dynamics have been discovered—for the last factor—there are no tests for new states of that factor. Similarly, when a new factor is added it becomes the last factor, and preceding factors are no longer eligible for new paths. In this setting, an observation comprises an epoch or sequence of outcomes that are necessary to infer state transitions. The examples in this paper used pairs of observations that enable the recognition of latent states and transitions from one latent state to another.

In principle, this kind of scheme should learn the structure of any (discrete) state-space model, given an appropriately ordered sequence of observations. However, there are some caveats here that rest upon the particular order in which observations are presented or ingested. In this sense, the structure learning is still supervised but in an implicit way; through the order in which training is structured over time. 

Specifically, the scheme above rests upon learning the likelihood mapping under precise beliefs about transitions. Similarly, learning the transition tensors rests upon precise likelihood mappings. This is assured when presenting training examples in the right order. In brief, the first path (of the last factor) is always known precisely because, by construction, it is a precise identity matrix. This means one has to present outcomes generated by all the states that constitute the factor in question. This is exactly the step illustrated in the previous section, when presenting styles of a given digit class. Because we are dealing with dynamics, this presentation has to be of stationary outcomes; e.g., a static visual image presented one or more times in succession. 

Once the requisite likelihood mapping has been learned, one can then present the different ways in which state-space can be traversed; e.g., a succession of previously seen words with a particular syntax. Or the presentation of an object, seen previously in distinct locations moving in a characteristic way. If these paths are trajectories that have been seen before, they will be explained by an existing path; otherwise a new path will be added. However, if we now introduce a new object or word, the most likely model will be one that is equipped with a new factor. Because this new factor has a precise path (a two-state identity matrix) the next observation will be assigned to the second state. It is assigned to the second state because the first state of the new factor is the same as the first state of discovered factors. This is necessary to ensure the factorial structure of the latent state-space is discovered. This means that whenever presenting states (and paths) for a new factor, these have to be presented under the first state (and path) of all preceding factors. Providing this curriculum is followed, the model will, in principle, automatically discover the right number of factors, and the right number of paths among the right number of states in each factor. 

Operationally, one can automatically generate the requisite curriculum of observations from a previously learned or specified generative model. These observations are then sufficient for the model to be learned from scratch; effectively duplicating a model via structure learning based solely on observations. However, as we will see in the next section, the learned model is not necessarily the same as the model generating outcomes because it is compressed in a Bayes optimal fashion. The annotated help for the (MATLAB) subroutine generating outcomes is as follows:

\begin{quote}
This routine generates a sequence of (probabilistic) outcomes from a specified POMDP structure. This sequence is appropriate for structure learning. It comprises a sequence of epochs, where each epoch is generated in a specific order: starting from the first factor, the outcomes associated with each hidden state are generated under the first path. By construction, the first path is stationary. After all hidden states have generated the outcomes, successive paths are generated starting from each hidden state. After all paths have been generated, the process is repeated for subsequent factors; under the first state and path of previous factors (noting, that the first path is always stationary; i.e., an identity transition mapping). Unless otherwise specified, outcomes comprise two observations.
\end{quote}

Anecdotally, this can be likened to teaching a child to read by presenting letters in isolation, so that they can be learned by populating likelihood tensors with appropriate Dirichlet counts. One can then move on to characteristic sequences; e.g., words. One can then repeat the procedure in another context; thereby, inducing a second factor; e.g., letters written in a different font—or spoken, as opposed to being read. 

This kind of learning is supervised in two senses. First, outcomes are presented in the right order for the generative model to assimilate them in a way that paths are learnable. In other words, a model would never learn the concept of motion if presented with randomly reordered video frames. Second, it is supervised because there is no action that selects the training samples. We will return to another aspect of ordinality later that rests upon being able to request or choose particular observations to ‘fill in’ missing parts of the likelihood tensor.

To illustrate the above scheme, we used a simple variant of the setup used to generate datasets such as dSprites; namely, different shaped objects moving—or movable—on a two-dimensional grid world \cite{champion2022branching, jeon2021ib}. To enable learning to be visualised in real time (in about 30 seconds using MATLAB on a personal computer), we used a simple setup in which the visual scene comprised 9×9 locations and objects were limited to 3 shapes that can move one pixel at a time in both dimensions. Following \cite{champion2022branching, fountas2020deep}, we introduced a reward or constraint by equipping the agents with an additional outcome modality; delivering reward when an object was in a certain location\footnote{The log prior preference for a reward present, over reward absent, was 4 natural units (i.e., a strong preference for sensing reward over no reward); i.e., about 50:1.}. Crucially, this location was different for each object. In effect, this creates a simple game for the agent: the agent has to recognise what it is looking at, and then move the object to its preferred location in a context sensitive fashion.

\begin{figure*}[t!]
    \centering
\includegraphics[width=0.42\textwidth]{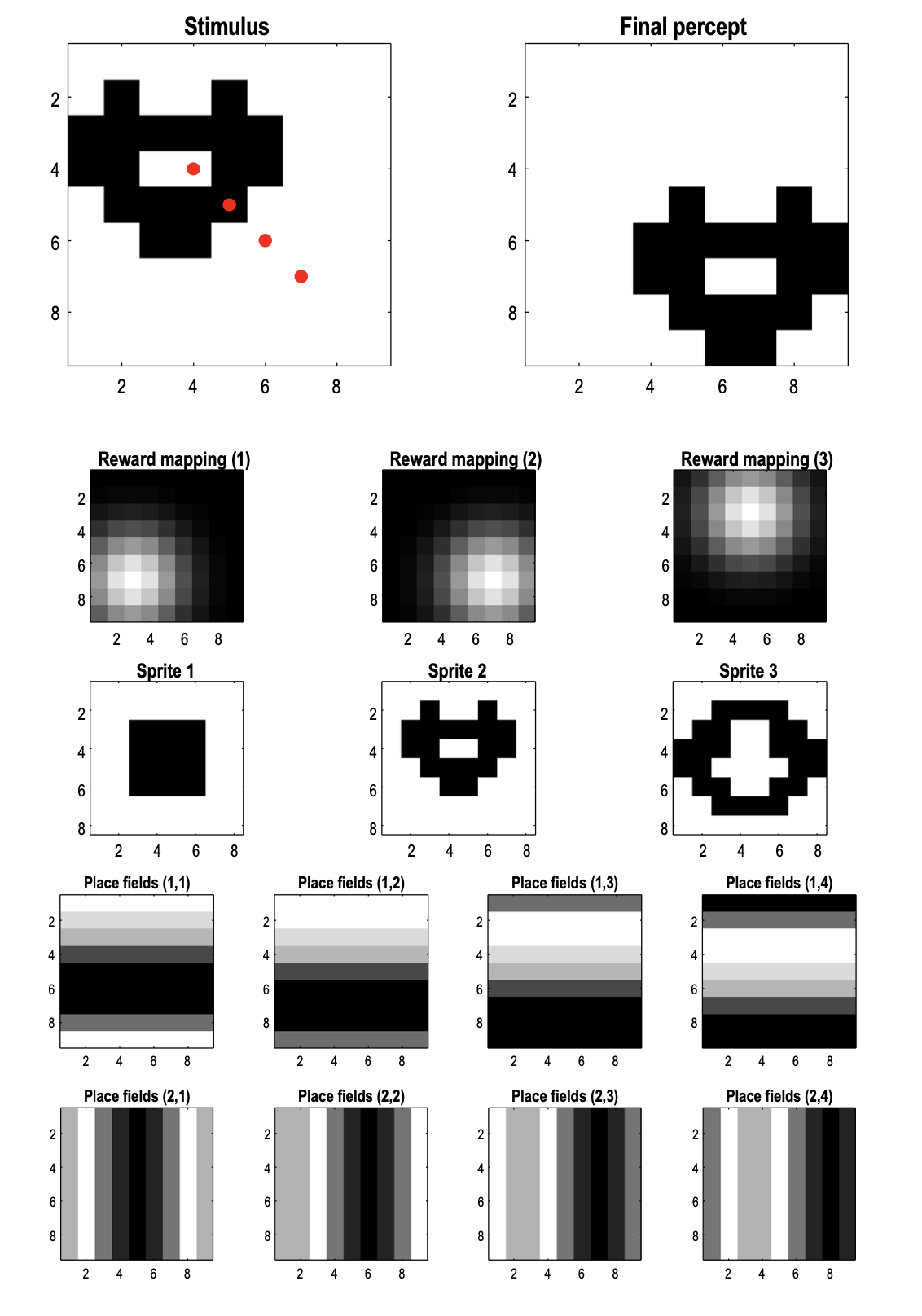}
\caption{\textbf{
From dSprites to Atari.
}  This figure illustrates the physics or game that the agent has to learn. In this world, there are three objects that can be moved up or down—or sideways—one pixel at a time. The agent has to learn the inherent physics of moving discrete objects, in a Euclidean space, which has two factors or dimensions; namely, horizontal and vertical movement. To showcase how the requisite structure learning underwrites agency, each object has to be moved to a unique location. The three objects are shown in the middle row, at the centre of their 9×9 pixel space. The preferred target locations for the three objects are shown in the second row, in the form of the probability of a reward outcome (a Gaussian function of distance from the preferred location). The structure of this world can be summarised as comprising 81 outcome factors or modalities of a visual sort, where each modality has two levels (\textit{black} or \textit{white}). In addition, there is a reward modality with two levels (\textit{absent} versus \textit{present}). These outcomes are generated by three factors. The first pair of factors corresponds to location in the \textit{horizontal} and \textit{vertical} dimensions, while the third factor generates the \textit{class} of the object and how it appears in outcome space. The lower rows show the representation of the location factors in the form of marginal densities over the likelihood mappings. In effect, these can be regarded as responses of ‘place cells’ that jointly specify the location of an object specified by the third factor. In this example—because of the circular boundary conditions—these ‘place fields’ are effectively periodic functions of vertical and horizontal locations, respectively. Neurobiologically, they can be thought of as a particular kind of place field featured by ‘boundary cells’ \cite{hartley2000modeling}. Equipped with this generative model, the agent can use prior preferences for sensing reward to infer plans of action and select the most likely movement of the object. The top two panels illustrate an example of this; moving the second object to its preferred location on the lower right. The red dots indicate the trajectory of the inferred—and realised—movement.
 }
\label{fig:figure_6}
\end{figure*}

To train an agent from scratch, we followed the above curriculum; presenting the first object in each of nine locations, starting from the first (upper left) location down to the lower left location. For each of these nine starting positions, we then presented transitions to the location above, and then the location below. This enables the agent to recognise a static object that can move up or down. We then repeated this for all horizontal locations—at the first vertical location—with the object moving in the left and right directions. For simplicity, we used circular boundary conditions, so that the objects appeared on the opposite side of the visual field when they moved across a boundary.
Figure \ref{fig:figure_6} illustrates the Bayes-optimal behaviour of an agent that has learned the structure of its sensed world. In this example, the agent recognises the object and that to get a reward it has to move the object to the lower left. It does this in the most efficient way possible by minimising expected free energy: see (\ref{eq:2}) and \cite{friston2021sophisticated}. This purposeful behaviour rests upon learning the structure of the generative model and the ensuing paths that can be controlled.

\begin{figure*}[t!]
    \centering
\includegraphics[width=0.7\textwidth]{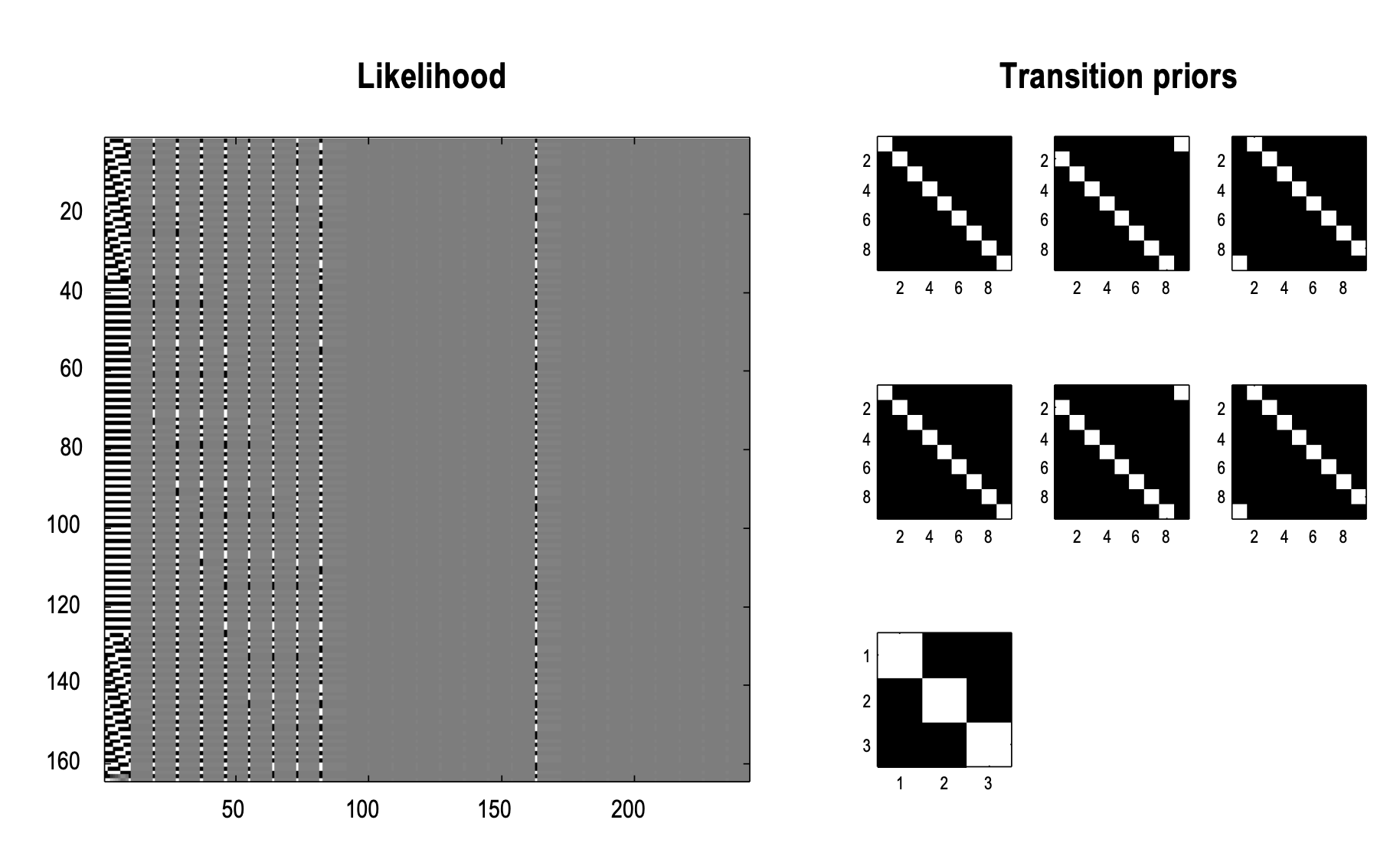}
\caption{\textbf{
Disentangling the structure of world models.}  This figure shows the structure of the likelihood and transition priors after exposure to a standard teaching sequence; namely, pairs of outcomes generated by states of the world and allowable transitions among states under the first state (and path) of previous factors. The left panel shows a concatenated likelihood tensor that has been unfolded to show the mapping between (9×9×3) states and the (9×9×2) + (1×2) outcomes. The transitions that have been discovered are shown on the right. Structure learning has discovered three factors; where the first pair of factors have three paths corresponding to no movement, moving in one direction or the other. The third factor has a single (stationary) path, which is an identity mapping. This implies that the object class is conditionally independent of motion. And motion in one dimension (i.e., factor) is conditionally independent of motion in the orthogonal dimension. In effect, the agent has disentangled the causes of observable outcomes; largely, in virtue of the factorisation afforded by the physics or dynamics of objects that move in a two-dimensional Euclidean space. Note that this generative model has no notion of metric space—other than implicit in the mappings entailed by the accumulated Dirichlet counts. Note further that the likelihood mapping is incomplete. There are many combinations of states that have yet to be experienced. Consequently, the Dirichlet counts of most columns in the likelihood mappings remain at their low initial values. It is these that have to be discovered via active learning to produce the ‘place field’ representations shown in the next figure.
 }
\label{fig:figure_7}
\end{figure*}

Figure \ref{fig:figure_7} illustrates the first stage of structure learning, showing the successive acquisition of transition tensors and their accompanying likelihood tensors. Although the stimuli were chosen carefully according to the protocol above, the agent just receives a stream of inputs and has to decide, with each successive input, whether to augment its model by adding states, paths, or new factors, as described above. The top row shows the discovered paths for the first (vertical location) factor. The first matrix is the first slice of the transition tensor and is always a precise identity matrix. However, the size of this matrix grows with the number of states. All subsequent stimuli can be explained in terms of these nine states and, after exposure to movement of the first object, the agent ‘knows’ that this object can either move up or down from every location. This is embodied by the two additional slices (i.e., matrices) of the transition tensors encoding movements or transitions from one location to the next. When the agent sees the first object in a new horizontal location, it induces a new factor and learns that there can be nine vertical locations—and that this object can move to the right or left. It retains its generative model until it encounters a new object that cannot be explained by any previously encountered latent state. At this point, it invokes a third factor with three levels, corresponding to the objects presented under the first state (and path) of the preceding objects. 

The agent now has the right structure and has discovered that there are three factors. The first two correspond to orthogonal motion or trajectories in some state-space, with nine levels each. The third factor provides the context for the object that moves in this way. However, the agent is far from having learned its generative model. It has never seen the second or third objects moving, or indeed, depart from their initial (first) states.

In more detail, although the model has a precise grip on the dynamics at hand (in the form of precise and complete transition tensors) there are many gaps in its experience that require more learning of the likelihood mapping. This is because the model has only been exposed to, literally, edge cases; in which each additional factor has been experienced under the first state (and path) of preceding factors. In other words, only certain edges of the likelihood tensor have accumulated Dirichlet parameters. For example, the agent has never seen the first object in the middle of the field of view. And has never seen the second object other than at the initial location. Filling in or accumulating model parameters would be a straightforward procedure if the model were equipped with precise posterior beliefs about latent states. However, to form precise posterior beliefs it has to learn the requisite likelihood mapping. So how can one elude this Catch-22? It is at this point that we turn to active inference and learning to ensure that the model accumulates the right kind of experience to fill in the gaps in its knowledge.

\subsection*{From models to agents}

This speaks to a second stage of \textit{self-supervised
} learning that rests upon equipping the model with agency by rendering certain paths controllable. In this instance, the paths correspond to motion in the horizontal and vertical dimensions, which the agent can control by switching between the paths that it has learned. To evince self-supervised learning (or autodidactic behaviour), we simply allowed the agent to select actions on the basis of expected free energy that, initially, is dominated by the novelty or expected information gain pertaining to the likelihood parameters: see (\ref{eq:2}). This means that the agent expects to move the objects into novel positions and cover the latent state-space as quickly and efficiently as possible. Because the agent has precise beliefs about state transitions—and can infer its initial states—it has precise beliefs about subsequent states and can therefore accumulate precise likelihood mappings given its observations. In short, the model is equipped with agency and can complete its acquisition of the likelihood tensor, for each of the three objects. 

\begin{figure*}[t!]
    \centering
\includegraphics[width=0.42\textwidth]{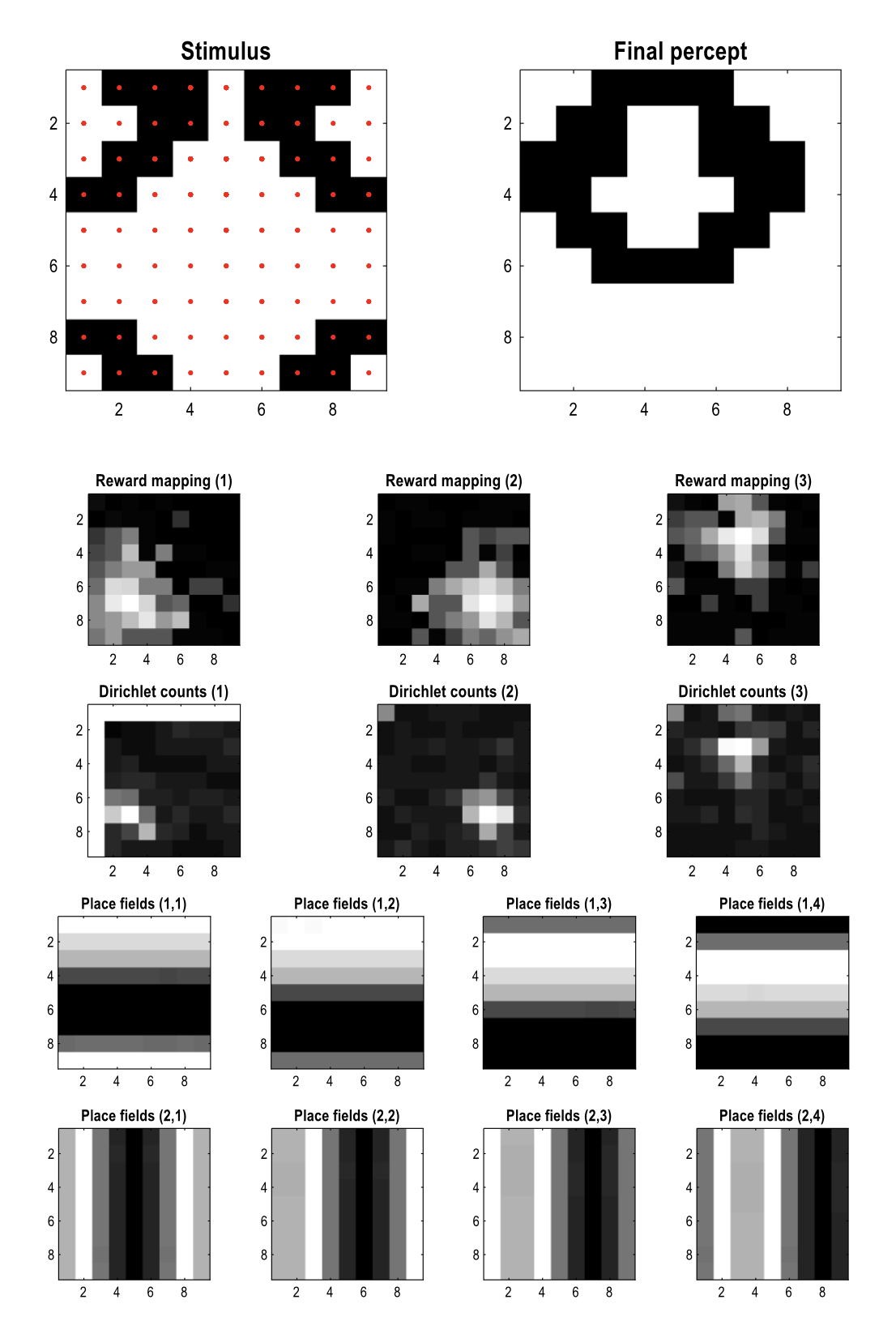}
\caption{\textbf{
Motor babbling.}  This figure illustrates explorative behaviour during active learning. The upper left panel shows the initial stimulus, while the upper right panel shows the final percept in its preferred location. The red dots indicate the locations visited over 512 epochs of a long explorative trial. Crucially, every location has been visited; in order to resolve uncertainty about the parameters of the likelihood mapping from the implicit latent state to outcomes. However, this behaviour is not purely explorative. This is because the agent was equipped with prior preferences that also have to be learned. This active learning is illustrated in the second row that shows the Dirichlet counts mapping from various locations to the reward outcome. The third row shows the total number of counts for each location. This scores the number of times each location is visited. It can be seen that for the first object, first horizontal location and first vertical location, the Dirichlet counts are very high (white). This is because these are the Dirichlet counts that were accumulated during structure learning, shown in Figure \ref{fig:figure_7}. The remaining locations, for each of the three objects, are those that have been actively explored and exploited during active learning. The accompanying receptive field (i.e., ‘place cell’) representation of these likelihood mappings are shown in the bottom two rows. These are virtually indistinguishable from the structure of the generative process in Figure \ref{fig:figure_6}. The evolution of various behaviours—and underlying belief distributions—are shown in the next figure.
 }
\label{fig:figure_8}
\end{figure*}

Figure \ref{fig:figure_8} shows the results of this active learning. The upper left panel illustrates the even coverage of locations (red dots) in the shortest amount of time. For example, by 128 trials, the agent had visited all possible locations at least once. The accompanying accumulation of Dirichlet counts is shown using the same format as Figure \ref{fig:figure_7}; in terms of the outcome modality reporting reward, and a series of marginal distributions that can be read as ‘place’ or ‘boundary cells’ from neurobiology \cite{hartley2000modeling}. A comparison of the implicit receptive fields in Figures \ref{fig:figure_6} and \ref{fig:figure_8} suggests that the agent has successfully learned the structure of its world model.

Figure \ref{fig:figure_9} shows the accompanying metrics of behaviour. Of note here, is the resolution of uncertainty about the likelihood mapping as scored by the reduction in (negative) expected free energy (i.e., expected information gain) after all locations have been visited (after about 81 trials). As the agent continues to familiarise itself with the causal structure of its sensorium—and the likelihood mappings become increasingly precise—the expected information gain or novelty falls until, after about 512 trials, the agent prefers to stay in its rewarding location, which it finds least surprising or costly.

From a neurobiological perspective, this kind of self-supervised learning can be associated with the explorative behaviour of infants, sometimes referred to as ‘motor babbling’ \cite{saegusa2009active}. When simulated in the context of developmental neurorobotics, the expected information gain is often referred to as the intrinsic motivation 
\cite{baranes2009r, barto2013novelty, oudeyer2007intrinsic, schmidhuber2010formal, schwartenbeck2019computational}.

\begin{figure*}[t!]
    \centering
\includegraphics[width=0.48\textwidth]{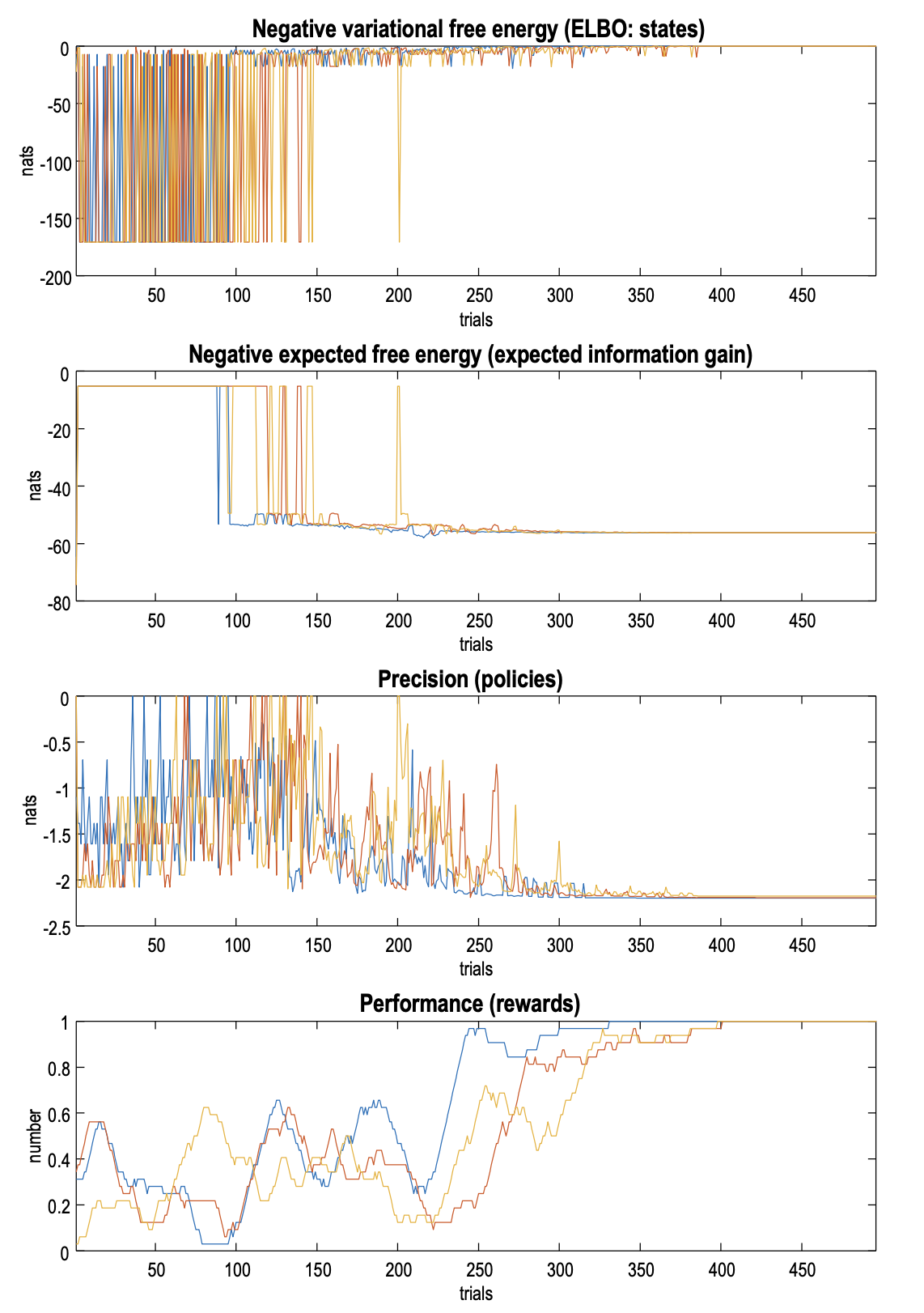}
\caption{\textbf{
Learning to plan.}  This figure reports the changes in various metrics as the agent becomes increasingly fluent with the structure of its world and engages in skilled, purposeful behaviour, specified in terms of prior preferences. The upper panel shows the (negative) variational free energy, which scores the log marginal likelihood of observations under the model, as it is learned over 512 epochs of one long trial or exposure. The second panel reports the expected information gain (averaged under posterior beliefs about action) as a function of trials. One can clearly see that after about 80 trials, the expected information gain falls dramatically as the novelty of revisiting each location is vitiated. At the same time, the (negative) variational free energy (i.e., marginal likelihood of observations) increases; in virtue of the fact that the agent can make more accurate predictions of what it will see. The third panel reports the precision over policies which, interestingly, increases during the initial exploration over the first hundred or so trials, during which there is a high expected information gain. Heuristically, this means that the agent is more confident about what it will do because it knows it can resolve uncertainty about “what it would see if it moved over there”. As the agent becomes more familiar and fluent with its environment, the remaining imperatives for precise action rest upon prior preferences. And the agent becomes progressively more exploitative as the expected cost falls below the expected information gain. This is shown in the final panel, in terms of performance; namely, the number of rewarding outcomes experienced. By about 300 trials, the agent just moves each of the three objects to their preferred locations—and rests thereafter. The three coloured lines correspond to the three objects in Figure \ref{fig:figure_6}.
}
\label{fig:figure_9}
\end{figure*}

\subsection*{Summary: the importance of being curious}

This treatment of (supervised) structure learning may seem a little contrived. However, it may be useful to consider why it works, to foreground the key mechanisms. To recap, by presenting stimuli or observations in a systematic order, we taught a generative model to assemble itself from scratch, such that it discovered latent factorial structure and dynamics. Furthermore, after this supervised structure learning, it was able to actively sample the world, to learn its causal structure with a few hundred observations. This is difficult or impossible to achieve using standard procedures in machine learning, such as manifold learning or nonparametric Bayes. The success of this model, based on implicit biomimetic imperatives, suggests that the structure learning problem is actually a structure teaching problem, in which agents can only learn if they are presented with material in the right way. Presenting material in the right way requires co-evolution of the teacher and learner of exactly the sort seen in evolutionary psychology and (cultural) niche construction 
\cite{constant2018variational, heyes2014cultural, laland1999evolutionary, shea2014supra}. If we read natural selection as nature's way of implementing Bayesian model selection 
\cite{campbell2016universal, frank2012natural, friston2023variational, vanchurin2022toward}, the emerging picture is a series of nested free energy minimising—i.e., evidence maximising—processes. On this view, protocols like the above teaching curriculum are themselves selected for structure learning via model selection. Active model selection enables Bayes-optimal active learning that, in turn, is rendered optimal by active planning as inference to maximise expected information gain—and, ultimately, optimise an agent’s grip on her world, through maximising the mutual information between observations and their latent causes. At the same time, all these nested processes supervene on each other. In the final section, we use the same procedures but in the context of problems usually addressed by providing an explicit description of the generative model, using things like Planning Domain Definition Languages \cite{fox2003pddl2}.

\section{Tower of Hanoi}\label{sec:tower_of_hanoi}

In the previous section we saw how structure learning underwrote the opportunity for the active learning of a generative model apt for disentangling the physics of a simple (e.g., Atari-like) game. Active learning in this example completed the likelihood mapping from latent states to observable outcomes; namely, the discrete (grey) levels of pixels. In this section, we apply the same procedures to the Tower of Hanoi problem \cite{donnarumma2016problem}. In this example, active learning is required to learn precise dynamics or transition probabilities among arrangements of balls or blocks. In the Tower of Hanoi problem, there are a small number of balls stacked on a small number of towers. The problem is to rearrange the balls into a target configuration, under the constraint that one can only move the top ball from one tower to another. This constraint stands in for the physics of the problem that has to be learned. After learning, one can then use active inference to solve the problem. 

As above, a sequence of outcomes were generated, where each sequence comprised a pair of outcomes. For simplicity, we considered three balls arranged among three towers. The outcomes in this instance were the states of each location— i.e., three levels of three towers—where each location could be one of three colours or empty. The set of ensuing (3×3) outcome modalities constitute a particular arrangement of the balls. The training sequence comprised all allowable arrangements, followed by all allowable transitions or ball rearrangements. The requisite sequence was generated by selecting one of the three balls and placing it on one of the three towers, selecting the next ball and placing it on every tower, and so on until all the balls were used up. This process was then repeated for every ordered sequence of coloured balls. 

\begin{figure*}[t!]
    \centering
\includegraphics[width=1\textwidth]{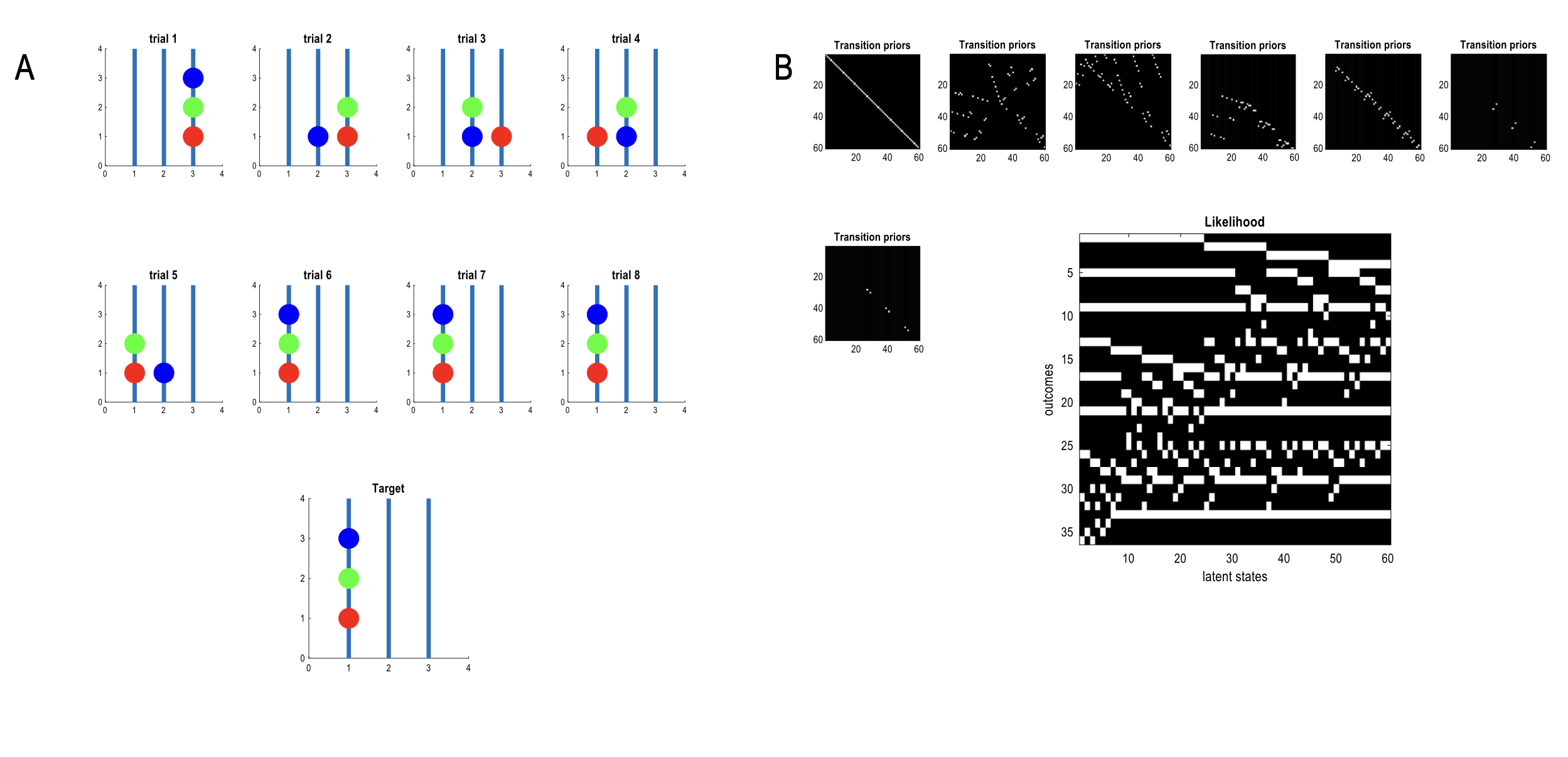}
\caption{\textbf{
Tower of Hanoi and its structure.}  \textbf{A}: this is a graphical illustration of the Tower of Hanoi problem with three balls arranged over three towers. The idea is to rearrange the balls into a target arrangement (lower panel), under the constraint that one cannot move a ball that is beneath another ball. In this example, it takes five moves to complete the problem. This was the solution found by an active inference agent that was equipped with the requisite likelihood mappings and constraints on transitions (i.e., rearrangements). Panel \textbf{B} shows the corresponding tensors following structure learning. The likelihood tensors have been concatenated and unfolded into a matrix; illustrating the mapping from 60 latent states (i.e., arrangements of the balls) to the outcomes. There are nine locations (i.e., modalities) with four outcomes each (one of three colours or empty) giving 36 distinct outcomes. In this example there is only one factor; namely, the arrangement of balls. The physics of rearranging the balls can be explained with seven paths, illustrated with slices of the transition tensor in the upper panels. Note that the first path is an identity mapping (i.e., no dynamics, by construction). Furthermore, note that some paths have indeterminate transition probabilities due to the unbalanced nature of the physics imposed by the constraints on moving balls. This ambiguity is resolved following active learning, as shown in the subsequent figures.
}
\label{fig:figure_10}
\end{figure*}

For every arrangement, the uppermost ball on every tower was moved to each of the three towers in turn. This training sequence clearly has some redundancy. For example, moving a ball to its own tower is the same as presenting the same arrangement twice. However, when ingesting this sequence\footnote{Using a low concentration parameter of 1/64, to install prior beliefs that likelihood and transition probabilities are sparse.}, the supervised structure learning automatically resolves this redundancy, explaining the entire sequence with 380 arrangements and seven paths (see Figure \ref{fig:figure_10}). As usual, the first path is stationary with the six remaining paths encoding an allowable rearrangement. However, as seen in Figure \ref{fig:figure_8}B, there are certain transitions from particular latent states (i.e., arrangements) that are not specified. This follows because of the redundancy above and the physics of this problem that inherits from the fact one cannot move a ball if it is beneath another ball. This physics does not lend itself to a simple factorisation or disentanglement of the sort seen in the previous section.

To resolve this imprecision or ambiguity, we then let the agent learn about the consequences of its action by rearranging the balls. In the absence of any prior preferences (e.g., for a target arrangement) active inference is driven purely by the expected information gain or epistemic affordances of moving the balls around.

\begin{figure*}[t!]
    \centering
\includegraphics[width=0.7\textwidth]{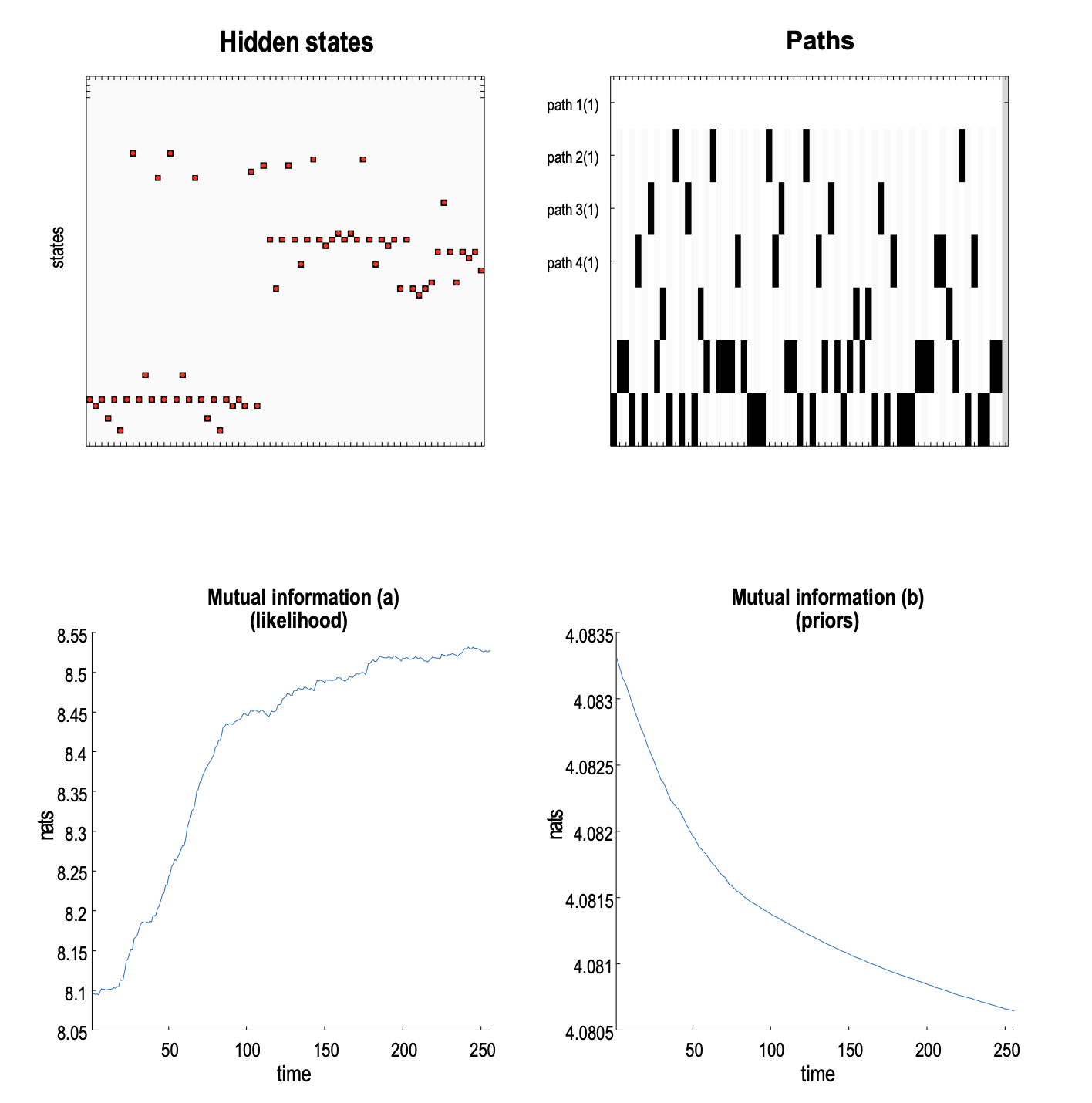}
\caption{\textbf{
Active learning.}  The upper panels show the states (left panel) and paths (right panel) chosen by an agent who could select one of the seven paths to move balls around. Actions were selected that minimise the expected free energy of predicted observations, where these predictions were furnished by active inference. The key thing to note here, is that the paths that have been explored more intensively are those which the agent is less certain about. This certainty is encoded by the accumulated Dirichlet counts under each path. For example, the last two paths in the previous figure are sparsely populated and therefore have the greatest epistemic affordance; and are therefore rehearsed more than the first path (about which the agent has very precise beliefs). The lower panels show the accumulation of mutual information over 256 time-steps, following the initial 64 time-steps illustrated in the upper panels.
}
\label{fig:figure_11}
\end{figure*}

Figure \ref{fig:figure_11} shows the ensuing behaviour in terms of the arrangements (states) during the first 64 moves of ‘motor babbling’. The states (and paths) visited demonstrate that the agent explored all paths with the exception of the first (stationary) path, about which it has precise knowledge. The information gain here pertains to the parameters of the generative model in (\ref{eq:2}); in particular, the prior transition parameters that compel the agent to explore moves or paths about which it has imprecise beliefs (i.e., low Dirichlet counts). The lower panels of Figure \ref{fig:figure_11} show the increase in mutual information of the likelihood mapping. The corresponding mutual information associated with the transition probabilities remains effectively the same (note the scaling of the Y axis). Active learning fills in the missing entries of the transition tensors as can be seen in panel A of Figure \ref{fig:figure_12} (compare this with the panel B of Figure \ref{fig:figure_10}). 

\subsection*{Planning and purpose}

\begin{figure*}[t!]
    \centering
\includegraphics[width=1.0\textwidth]{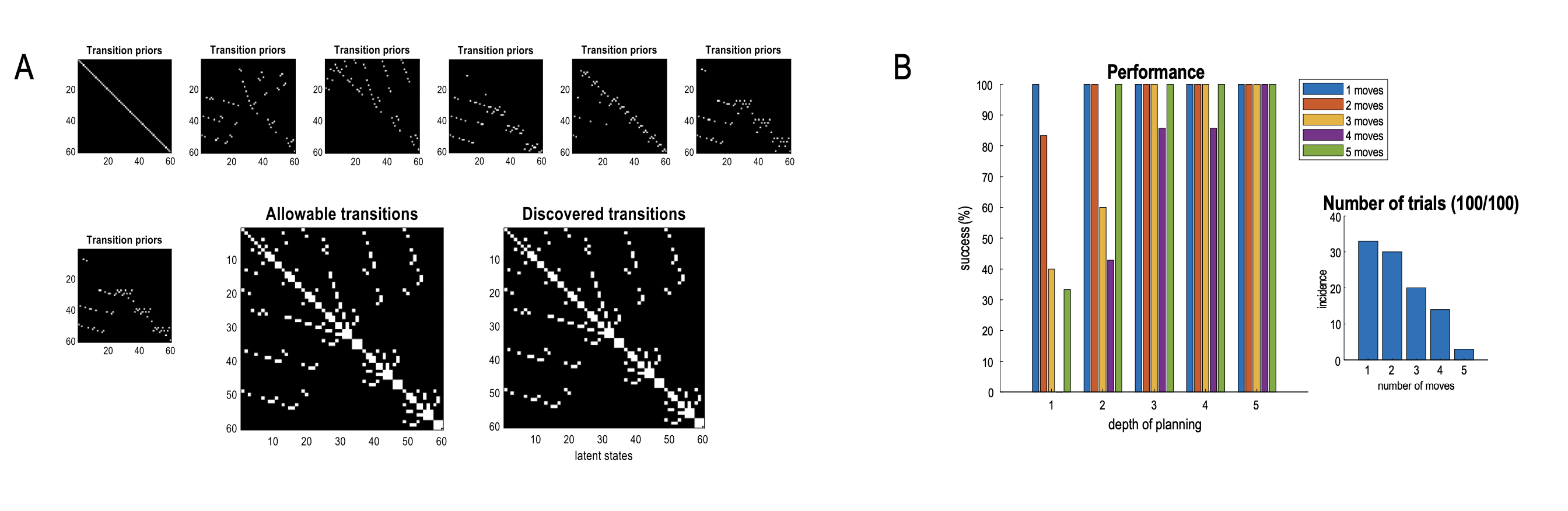}
\caption{\textbf{
Discovery and performance.} Panel \textbf{A} shows the transition tensors after active learning. Note that the sparsely populated slices of the transition tensor after structure learning (\ref{fig:figure_10}) have been filled in by active learning. The lower panels show the marginal distribution of transitions over paths to illustrate that all allowable transitions have been discovered. This discovery enables the agent to plan trajectories into the future and—if equipped with prior preferences—rearrange the balls into some preferred arrangement. Panel \textbf{B} illustrates the ensuing performance. The right panel illustrates the proportion of 100 trials that were considered easy, through to hard (requiring 1 to 5 moves, respectively). The accompanying performances are shown in the left panel for agents with an increasing depth of planning (from 1 to 5 moves). The takeaway from these results is that as the depth of planning approaches the number of moves required to solve the problem, performance increases to 100\%. For example, for trials that require 4 moves, agents with a planning depth of two only attain about a 40\% success rate. Increasing the planning depth to 4 increases performance to about 85\%, while a planning depth of 5 ensures 100\% performance.
}
\label{fig:figure_12}
\end{figure*}

To illustrate the purposeful or goal-directed behaviour that can now be elicited—by equipping the agent with prior preferences—we simulated 100 trials, each comprising 8 moves. On each trial a target arrangement was specified in terms of preferences over outcomes; i.e., each of the nine locations\footnote{Here, the preferences for observing the target location were fairly mild with a difference between the preferred and nonpreferred outcomes of one natural unit: i.e., about 3:1.}. The targets were chosen at random to cover easy (one move) and hard (five move) problems. The agent was able to complete the task in 100\% of the trials; provided it was equipped with a sufficient depth of planning; here, five steps into the future \cite{friston2021sophisticated}. Panel B in Figure \ref{fig:figure_12} shows the distribution of trials with different numbers of moves—and the percentage success—as a function of the depth of planning. The takeaway from these results is that as the depth of planning increases, the frequency of solving more difficult problems increases until 100\% performance is attained when the planning depth is sufficiently large. These results were obtained by repeating every trial, using the same initial and target arrangements, for an agent with planning depth of one through to five.

\subsection*{Summary: the importance of thinking ahead}

In summary, this section demonstrates the generality of supervised structure learning for ingesting the right kind of training observations and using the ensuing structure to scaffold the active learning of requisite model parameters. This active learning is driven by expected information gain that is one facet of minimising expected free energy (or maximising mutual information). With the right generative model, planning and preference-seeking behaviours are straightforward to elicit by equipping the agent with some prior preferences or constraints on the outcomes it expects to encounter. 

The above simulations in MATLAB run in minutes on a personal computer. This efficiency—in terms of compute time—is a signature of active inference in the sense that everything rests on minimising path integrals of variational and expected free energy—and, implicitly, minimising complexity cost and risk, respectively (i.e., the degree of Bayesian belief updating and attendant computations).

The latitude to train generative models and leverage their auto-didactic capabilities, within a minute or so, speaks to the possibility of a single use or disposable (generative) AI. In other words, instead of training an overly expressive neural network on large cohorts of data—for every plausible application context—it is, in principle, possible to create and deploy generative models in a few minutes that have the agency requisite for solving a particular problem in a particular context. In one sense, this is how we operate; e.g., when solving a crossword puzzle, whose structure has never been seen before, and will never be seen again.

\section{Discussion}\label{sec:discussion}

The approaches illustrated by the above numerical studies speak to three key themes: namely, inference versus learning; discrete versus continuous; and teaching versus learning. We conclude with a brief discussion along these lines and a comparison of the relative merits of committing to one approach or the other.

\subsection*{Inference versus learning}

A fundament of active inference—and the free energy principle from which it inherits—is that everything can be cast as an optimisation process, where variational free energy is minimised (or marginal likelihood is maximised) over nested timescales. On this view, model selection is a slow process that depends upon optimising the parameters of generative models. Optimising the parameters of generative models depends upon inferring the latent states generating content. In short, the same belief updating processes are nested so that each informs—and is informed by—the same kind of process at faster and slower timescales. This is mandated in variational Bayes by the requisite message passing between the factors of a mean field approximation. In turn, this has the implication that learning is a belief updating process that requires the encoding of posterior distributions over model parameters.

Interestingly, the majority of machine learning approaches do not estimate uncertainty about parameters; even in state-of-the-art variational autoencoders and current implementations of predictive coding \cite{luo2022understanding, marino2022predictive, mescheder2017adversarial}. The fully (variational) Bayesian approach considered here offers a perspective on learning, not as optimising the weights or parameters of a neural network to optimise recognition, classification or inference, but rather as a process of finding the most likely distribution over model parameters. If one subscribes to this view, it means that machine learning applications that do not accommodate posteriors over parameters are, effectively, learning to infer, via amortisation \cite{mazzaglia2022free, zhang2018advances}. 

This contrasts with the learning illustrated above in which beliefs about model parameters were optimised with respect to an ELBO before evaluating the marginal likelihood of each successive data point. On this view, learning becomes a kind of inference that has its own optimality criteria—enabling one-shot learning and a Bayes-optimal assimilation or ingestion of data. This is evidenced in the above simulations, in which the number of training exemplars is an order of magnitude smaller than typically used in machine learning of a more conventional sort. Similar arguments can be made at the next level of belief updating; namely, Bayesian model selection. In order to select the most likely model, one has to evaluate the evidence for that model. To do so, one has to marginalise over uncertainty about model parameters. Again, this speaks to the importance of encoding beliefs or uncertainty about model parameters to realise the kind of optimality on offer from variational (i.e., approximate) Bayesian inference. If this argumentation is correct, it motivates the inclusion of densities over model parameters in conventional approaches; for example, Gaussian distributions over continuous state-space models or Dirichlet distributions over discrete state-space models. In many senses, this has already been addressed outside machine learning and is a commonplace procedure in complex and dynamical systems modelling.

\subsection*{Discrete versus continuous}

The illustrations above commit to a discrete state-space model. This was largely to finesse the computational complexity of evaluating requisite free energy functionals. In discrete state-spaces, these evaluations rest largely upon sum-product operators on tensors. As noted above, this eschews worries about nonlinear activation functions or the composition of such nonlinear mappings in deep neural network architectures. It also eludes constraints of differentiability, via the chain rule, upon which procedures like backpropagation rest. Having said this, at the end of the day, the imperative for any model is to minimise complexity while maintaining a high predictive accuracy. Complexity here is the information gain or KL divergence between posterior and prior beliefs that is reflected in the number of model parameters. Discrete state-space models effectively trade analytic complexity for high dimensional state-spaces and, implicitly, more model (e.g., Dirichlet) parameters. 

The foregoing suggests that a first principles argument can be made for continuous state-space models if the number of parameters can be substantially reduced. There are lots of nice examples of this in the machine learning literature, such as the weight sharing implicit in convolutional neural networks, which implicitly embody a prior over the contiguity properties of the data or content generating process (e.g., for images). One could argue that an advantage of discrete state-space models is that one can optimally coarse-grain discrete state-spaces to provide an apt explanation for any particular kind of content. However, ultimately, the question of whether to use continuous versus discrete state-space models rests upon the respective model evidence. In turn, this suggests that it might be useful to extend the variational procedures illustrated above to continuous state-space models with beliefs over states and parameters (and precisions). Most of the analytic results are at hand with one exception: namely, the evaluation of expected free energy over continuous trajectories into the future. In principle, this should succumb to a path integral formulation but, to our knowledge, has not yet been addressed fully: but see \cite{ccatal2019bayesian}.

One of the limitations of discrete state-space models is that data are ingested or assimilated rather slowly. Although the structure learning of the dynamics in the second and third examples run in real-time (a minute or so), the MNIST example can take about an hour to assimilate 10,000 digits. This is because it is effectively inferring the cause of each digit and updating (Dirichlet) parameters, for each observation. This can take a few hundred milliseconds for high-dimensional problems (e.g., images).

\subsection*{Teaching versus learning}

If one reads structure learning as Bayesian model selection, there are basically two approaches. One can start with an overly expressive or over-parameterised model and remove redundant components by comparing the evidence for the full or parent model and the reduced model. This Bayesian model reduction is used extensively for learning structures and architectures in complex system modelling. The second approach is to grow a model; a little bit like growing a crystal, by adding material if it minimises free energy (i.e., increases model evidence). The particular simplification of working with discrete state-space models is that the directions of growth are well-defined—and can be explored in a straightforward way, as above. 

From a biomimetic perspective, this foregrounds the importance of acquiring new knowledge via experience-dependent learning where, crucially, that experience is consilient with the way structured knowledge is accumulated. This is the sense in which teaching is used in the current setting. In theoretical biology, this rests upon cultural niche construction and, in an educational setting, curriculum learning. The curricula at hand rest upon presenting content or experiences in the right order, to ensure that learning the likelihood mappings precedes the learning of transition priors—and learned transition priors enable likelihood learning; where the implicit bootstrapping may rest upon active learning. This enactive component again has a biomimetic aspect, in the sense that to teach efficiently requires a curious student who, at some point, can become autodidactic. The implicit kind of supervised structure learning raises a question: where do the requisite curricula come from?

One could argue that the process generating content offers the right curriculum; simply because it has dynamics. In other words, all that is required is that things are first seen (or fixated) for a sufficient period of time to accumulate Dirichlet parameters in a likelihood mapping. Following this, any characteristic dynamics or movements can be explained by learning transition tensors, until a new kind of object is encountered, and the process starts again. Clearly, there are additional constraints on the order of events implicit in the curricula used in the numerical studies (e.g., all possible movements have to be presented before moving to the next object or context and, in factorial situations, there are constraints on initial conditions); however, a lot of the requisite ordinality exists in the temporal structure of any process generating content.

While we have demonstrated the benefits of a curriculum in structure learning, this leads to another interesting question: what is the most effective curriculum? To be able to answer this, it is first necessary to think about the degrees of freedom in designing curricula. As an example, consider extending our MNIST example to learning to read handwriting. Here, as in the Atari-like problem above, we must account for dynamics in predicting the next letters or words in a sequence (of fixations). We could select a curriculum in which, as above, we first learn the mapping from letter states to all the ways in which they can be written, and then learn about the (transition) statistics of sequences of letters in written language. Alternatively, we could start with a set of Platonic letters and learn about the paths they are likely to follow, and then use these paths as informative priors to assist in learning about handwriting styles (i.e., learning the likelihood tensor structure). The latter might be more like studying a map or an abstract schematic before exploring or studying a real system. In principle, it may be possible to treat different curricular permutations as alternative policies and to assess these through their associated expected free energy, and this may be an interesting direction for future research.

In concluding, we emphasise the importance of temporality or ordinality. This is because it licences the structure learning of deep or hierarchical models with separation of temporal scales. Recall that the \textit{raison d'être} for supraordinate levels is to model transitions from one context to the next, where the context specifies the initial conditions (and path) for the level below. Practically, this means that one could (after a replay or backward pass) present a sequence of initial states (or paths) to a new model and repeat the procedure outlined above. This new model would learn slow contextual structure that provides inductive biases (i.e., empirical priors) that inform—and are informed by—subordinate levels. For example, we could create a deep generative model in the dSprites demonstration and test for any Markovian aspect to the order in which objects appeared; thereby endowing the generative model with a deep, semi-Markovian, context sensitivity.

\paragraph{Acknowledgements}
KF is supported by funding for the Wellcome Centre for Human Neuroimaging (Ref: 205103/Z/16/Z), a Canada-UK Artificial Intelligence Initiative (Ref: ES/T01279X/1) and the European Union’s Horizon 2020 Framework Programme for Research and Innovation under the Specific Grant Agreement No. 945539 (Human Brain Project SGA3). LD is supported by the Fonds National de la Recherche, Luxembourg (Project code: 13568875). This publication is based on work partially supported by the EPSRC Centre for Doctoral Training in Mathematics of Random Systems: Analysis, Modelling and Simulation, United Kingdom (EP/S023925/1). 

\section*{Disclosure statement}\label{sec:disclosure_statement}
The authors have no disclosures or conflict of interest.

\appendix

\section{Universal Generative Models}
\label{app:ugm}

The generative model presented in Figure~\ref{fig:figure_1} can be seen as a ``universal generative model'', as it provides a general treatment for modelling any agent in any environment. Moreover, the same model structure can be stacked hierarchically, where a higher level model contextualises the prior over states and paths (through D and E) of a lower level model. The lower level in turn gathers evidence for the higher level state inference. This provides a scale-free architecture where each higher level operates at a coarser timescale, facilitating long time horizon planning and inference. Within each hierarchical level, the state space can also be further factorized, endowing the model with factorial depth. In essence, the model hence `carves nature at its joints' into factors that interact to generate outcomes. The resulting classes of universal generative models are depicted in Figure~\ref{fig:figure_1}.

\begin{figure*}[b!]
    \centering
    \includegraphics[width=1.0\textwidth]{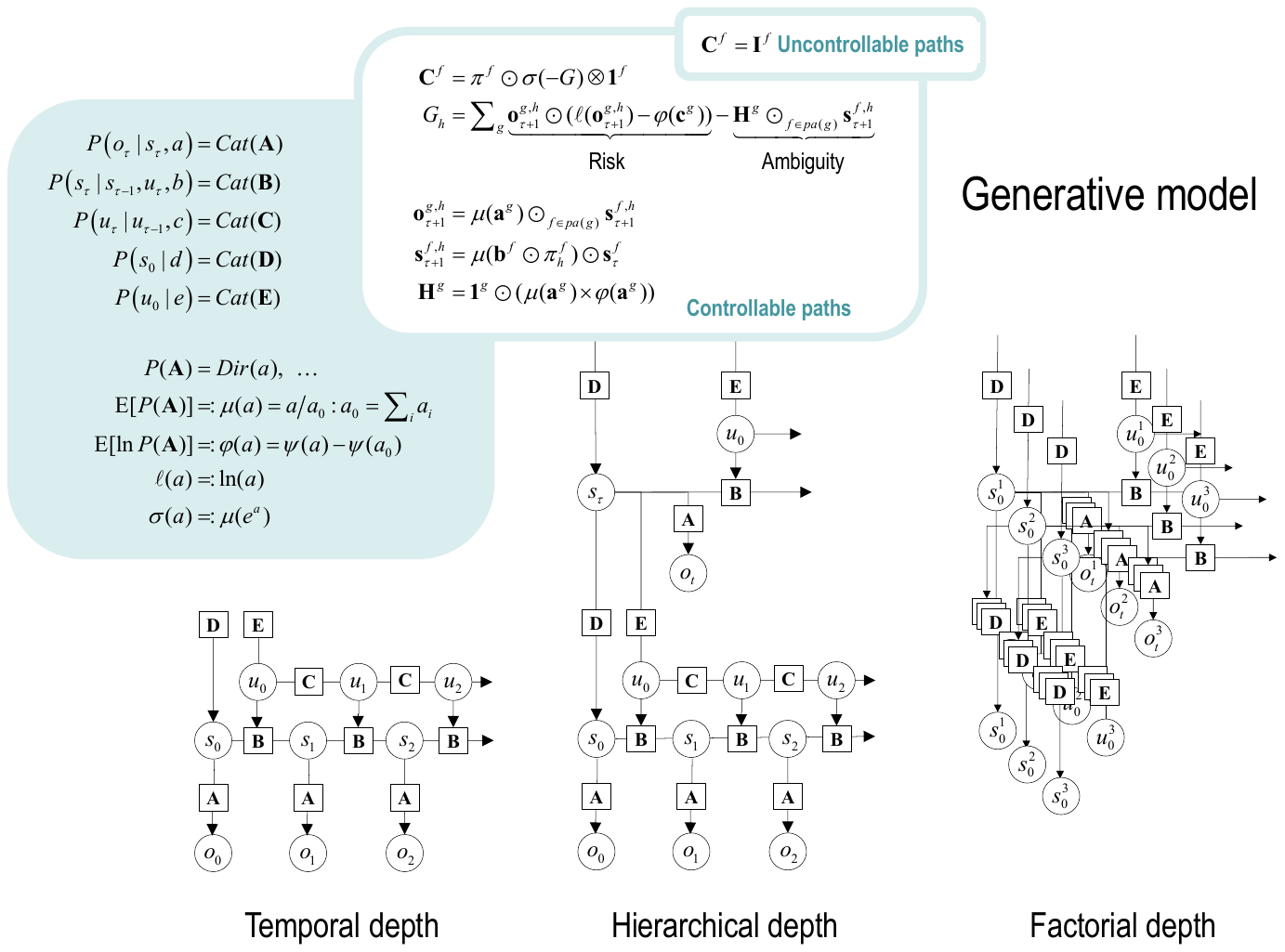}
\end{figure*}
\begin{figure*}
    \caption{\textbf{Generative models as agents.} A generative model specifies the joint probability of observable consequences and their hidden causes. Usually, the model is expressed in terms of a \textit{likelihood} (the probability of consequences given their causes) and \textit{priors} (over causes). When a prior depends upon a random variable it is called an \textit{empirical prior}. Here, the likelihood is specified by a tensor $\mathbf{A}$, encoding the probability of an outcome under every combination of \textit{states} ($s$). The empirical priors pertain to transitions among hidden states, $\mathbf{B}$, that depend upon \textit{paths} ($u$), whose transition probabilities are encoded in $\mathbf{C}$. Certain (controllable) paths are more probable \textit{a priori} if they minimise their expected free energy ($\mathbf{G}$), expressed in terms of \textit{risk} and \textit{ambiguity} (lower white panel). If the path is not controllable, it remains unchanged during the epoch in question (upper white panel), where $\mathbf{E}$ specifies the empirical prior probability of each path. The left panel provides the functional form of the generative model in terms of categorical ($Cat$) distributions. The lower equalities list the various operators required for the variational message-passing detailed in Figure \ref{fig:figure_2}. These functions are taken to operate on each column of their tensor arguments. The graph on the lower left depicts the generative model as a probabilistic graphical model that foregrounds the implicit \textit{temporal depth} implied by priors over state transitions and paths. This example only shows dependencies for uncontrollable paths. When equipped with \textit{hierarchical depth} the POMDP acquires a separation of temporal scales.  This follows from a construction in which the succession of states at a higher level generates the initial states (via the $\mathbf{D}$ tensor) and paths (via the $\mathbf{E}$ tensor) at the lower level. This means higher levels unfold more slowly than lower levels, thereby furnishing empirical priors that contextualise the dynamics of lower level states. At each hierarchical level, hidden states and accompanying paths are factored to endow the model with \textit{factorial} depth. In other words, the model or agent ‘carves nature at its joints’ into factors that interact to generate outcomes (or the initial states and paths at lower levels). This means context-sensitive contingencies are mediated by the tensors mapping from one level to the next ($\mathbf{D}$ and $\mathbf{E}$) or outcomes ($\mathbf{A}$). The subscripts in this figure pertain to time, while the superscripts denote different factors ($f$), outcome modalities ($g$) and combinations of paths over factors ($h$). Tensors and matrices are denoted by uppercase bold, while posterior expectations are in lowercase bold. The matrix $\pi$ encodes the probability over paths, under each policy (where $\pi_h$ denotes the probability over paths for policy $h$). The $\times$ notation implies a generalised inner (i.e., dot) product or tensor contraction, while $\odot$ denotes the Hadamard (element by element) product. $ch(\cdot)$ and $pa(\cdot)$ return the children and parents of any node; namely, the co-domain and domain, respectively. $\psi(\cdot)$ is digamma function, applied to the columns of a tensor.}
\label{fig:figure_1}
\end{figure*}
\newpage

\subsection*{Belief updating}

\begin{figure*}[t!]
    \centering
    \includegraphics[width=1.0\textwidth]{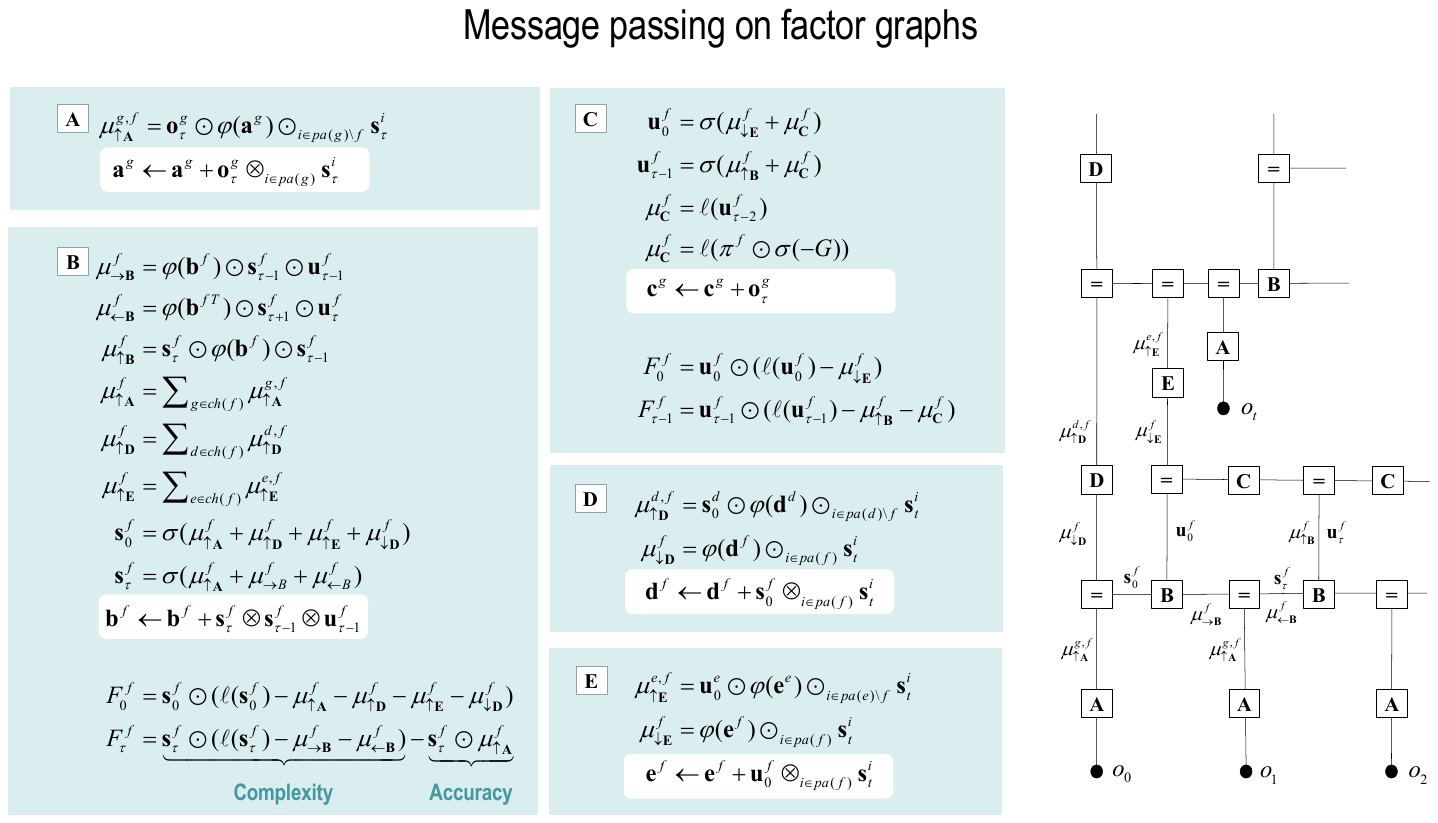}
\caption{\textbf{Belief updating and variational message passing: } the right panel presents the generative model as a factor graph, where the nodes (square boxes) correspond to the factors of the generative model (labelled with the associated tensors). The edges connect factors that share dependencies on random variables. The leaves of (filled circles) correspond to known variables, such as observations ($o$). This representation is useful because it scaffolds the message passing—over the edges of the factor graph—that underwrite inference and planning. The functional forms of these messages are shown in the left-hand panels. For example, the expected path—in the first equality of panel \textbf{C}—is a softmax function of two messages. The first is a descending message $\mu_{\downarrow_{\mathbf{E}}}^f$ from $\mathbf{E}$ that inherits from expectations about hidden states at the hierarchical level above. The second is the log-likelihood of the path based upon expected free energy. This message depends upon Dirichlet counts scoring preferred outcomes—in modality $g$---encoded in $\mathbf{c}^{g}$ (see Figure \ref{fig:figure_1}). The two expressions for $\mu_{\mathbf{C}}^f$ correspond to uncontrolled and controlled paths, respectively. The updates in the lighter panels correspond to learning; i.e., updating Bayesian beliefs about parameters (adopting the Einstein summation with respect to $\tau$). Similar functional forms for other
messages can be derived, by direct calculation. The $\odot$  notation implies a generalised inner product or tensor contraction, while $\otimes$ denotes an outer product. $ch(\cdot)$ and $pa(\cdot)$ return the children and parents of any node; namely, the domain and co-domain, respectively. There is a link (i.e., edge) between a parent and child, if the codomain of the parent factor (i.e., node) constitutes a domain of a child. This means that each edge—and (the codomain of) each factor—is uniquely associated with a state, path or outcome. }
\label{fig:figure_2}
\end{figure*}

In variational treatments, the sufficient statistics encoding posterior expectations are updated by minimising variational free energy. Figure \ref{fig:figure_2} illustrates these updates in the form of variational message passing \cite{dauwels2007variational, friston2017graphical, winn2005variational} for a universal generative model. For example, expectations about hidden states are a softmax function of messages that are linear combinations of other expectations and observations.

\begin{equation}
\begin{aligned}
\mathbf{s}_\tau^f & =\sigma\left(\mu_{\uparrow_{\mathbf{A}}}^f+\mu_{\rightarrow B}^f+\mu_{\leftarrow B}^f\right) \\
\mu_{\uparrow_{\mathbf{A}}}^f & =\sum_{g \in c h(f)} \mu_{\uparrow \mathbf{A}}^{g, f} \\
\mu_{\uparrow_{\mathbf{A}}}^{g, f} & =\mathbf{o}_\tau^g \odot \varphi\left(\mathbf{a}^g\right) \odot_{i \in p a(g) \backslash f} \mathbf{s}_\tau^i
\label{eq:4}
\end{aligned}
\end{equation}

Here, the ascending messages from the likelihood factor are a linear mixture\footnote{The notation implies a sum product operator; i.e., the dot or inner product that sums over one dimension of a numeric array or tensor. In this paper, these sum product operators are applied to a vector $\mathbf{a}$ and a tensor $\mathbf{A}$ where, $\mathbf{a} \odot \mathbf{A}$ implies the sum of products is taken over the leading dimension, while $\mathbf{A} \odot \mathbf{a}$ implies the sum is taken over a trailing dimension. For example, $\mathbf{1} \odot \mathbf{A}$ is the sum over columns and $\mathbf{A} \odot \mathbf{1}$ is the sum over rows, where $\mathbf{1}$ is a vector of ones. This notation replaces the Einstein summation notation to avoid visual clutter.} of expected states and observations, weighted by (digamma) functions of the Dirichlet counts that correspond to the parameters of the likelihood model (c.f., connections weights). The expressions in Figure \ref{fig:figure_2} are effectively the fixed points (i.e., minima) of variational free energy. This means that message passing corresponds to a fixed point iteration scheme that inherits the same convergence proofs of coordinate descent \cite{beal2003variational, dauwels2007variational, winn2005variational}. Under these—discrete state-space—generative models, message passing generally converges within a couple of iterations (and in one iteration when only passing forward messages during online Bayesian filtering).

\begin{sloppypar}
\printbibliography[title={References}]
\end{sloppypar}

\end{document}